\documentclass[journal]{IEEEtran}

\usepackage{array}
\usepackage{cite}
\usepackage{mdwmath}
\usepackage{graphicx}        
\usepackage{subfigure}
\usepackage{pifont}

\usepackage{bbm}
\usepackage{dsfont}

\usepackage{bm}
\usepackage{comment}
\usepackage{amsmath}
\usepackage{enumerate} 
\usepackage{mathrsfs}
\usepackage{makeidx}         
\usepackage{graphicx}       
\usepackage{subfigure}
\usepackage{epsfig,amssymb,latexsym}
\usepackage{psfrag}
\usepackage{algorithm}
\usepackage{algorithmicx}
\usepackage{subfigure}
\usepackage{algcompatible,lipsum}
\usepackage{cases}
\usepackage[misc]{ifsym}
\usepackage{fancyhdr}
\usepackage{multirow}
\usepackage{color}
\usepackage{indentfirst}
\usepackage{hyperref}
\usepackage{colortbl}
\definecolor{lightgray}{rgb}{.93,.93,.93}

\newcommand{\mr}{\mathrm}

\begin{document}

\title{InOR-Net: Incremental 3D Object Recognition Network for Point Cloud Representation} 

\author{
	Jiahua Dong,
	Yang Cong, \IEEEmembership{Senior Member,~IEEE,}
	Gan Sun, \IEEEmembership{Member,~IEEE,}
	Lixu Wang,
	Lingjuan Lyu, \\
	Jun Li, 
	and 
	Ender Konukoglu
	
	\thanks{This work was supported in part by the National Nature Science Foundation of China under Grant 62127807, 62225310, 62273333, 62133005; and in part by the State Key Laboratory of Robotics under Grant 2023-Z13. (\textit{Corresponding author: Yang Cong.}) }
	
	\thanks{Jiahua Dong is with the State Key Laboratory of Robotics, Shenyang Institute of Automation, Chinese Academy of Sciences, Shenyang 110016, China, and also with the University of Chinese Academy of Sciences, Beijing 100049, China (e-mail: dongjiahua1995@gmail.com).}
	
	\thanks{Yang Cong and Gan Sun are with the State Key Laboratory of Robotics, Shenyang Institute of Automation, Chinese Academy of Sciences, Shenyang 110016, China (e-mail: congyang81@gmail.com, sungan1412@gmail.com).}	
	
	\thanks{Lixu Wang is with the Computer Science Department, Northwestern University, Evanston, USA (e-mail: lixuwang2025@u.northwestern.edu).}
	
	\thanks{Lingjuan Lyu is with the Sony AI, Tokyo 108-0075, Japan (e-mail: Lingjuan.Lv@sony.com).}
	
	\thanks{Jun Li is with the School of Computer Science and Engineering, Nanjing University of Science and Technology, Nanjing 210094, China (e-mail: junli@njust.edu.cn).}	 
	
	\thanks{Ender Konukoglu is with the Computer Vision Lab, ETH Zurich, Z\"{u}rich 8092, Switzerland. (e-mail: ender.konukoglu@vision.ee.ethz.ch). }
}

\markboth{IEEE Transactions on Neural Networks and Learning Systems,~Vol.~14, No.~8, March~2023}%
{Shell \MakeLowercase{\textit{et al.}}: Bare Demo of IEEEtran.cls for IEEE Journals}

\maketitle

\begin{abstract}
3D object recognition has successfully become an appealing research topic in the real-world. However, most existing recognition models unreasonably assume that the categories of 3D objects cannot change over time in the real-world. This unrealistic assumption may result in significant performance degradation for them to learn new classes of 3D objects consecutively, due to the catastrophic forgetting on old learned classes. Moreover, they cannot explore which 3D geometric characteristics are essential to alleviate the catastrophic forgetting on old classes of 3D objects. To tackle the above challenges, we develop a novel \underline{In}cremental 3D \underline{O}bject \underline{R}ecognition \underline{Net}work (\emph{i.e.}, InOR-Net), which could recognize new classes of 3D objects continuously via overcoming the catastrophic forgetting on old classes. Specifically, a category-guided geometric reasoning is proposed to reason local geometric structures with distinctive 3D characteristics of each class by leveraging intrinsic category information. We then propose a novel critic-induced geometric attention mechanism to distinguish which 3D geometric characteristics within each class are beneficial to overcome the catastrophic forgetting on old classes of 3D objects, while preventing the negative influence of useless 3D characteristics. In addition, a dual adaptive fairness compensations strategy is designed to overcome the forgetting brought by class imbalance, by compensating biased weights and predictions of the classifier. Comparison experiments verify the state-of-the-art performance of the proposed InOR-Net model on several public point cloud datasets. 

\end{abstract}


 \ifCLASSOPTIONpeerreview
 \begin{center} \bfseries EDICS Category: 3-BBND \end{center}
 \fi
%
\IEEEpeerreviewmaketitle

\section{Introduction}
\IEEEPARstart{O}{bject} recognition technology in 3D vision has attracted widespread attention in the various research fields. Until now, it has achieved great successes in many real-world challenging applications including intelligent robotics \cite{8593741}, environmental surveillance \cite{Roynard_2018_CVPR_Workshops}, autonomous driving \cite{Behl_2017_ICCV}, medical diagnosis \cite{What_Transferred_Dong_CVPR2020}, etc. To better characterize the semantic context of unordered 3D point cloud data collected by the LiDAR system or depth camera, plenty of convolutional neural networks \cite{Qi_2017_CVPR, NIPS2017_7095} are introduced to capture discriminative 3D geometric properties and explore the task-specific representations. These convolutional architectures \cite{NIPS2018_7362, Lan_2019_CVPR, yan2020pointasnl, yu2022generalized} significantly improve the characterization ability on the task of 3D object recognition for point cloud representation, by encoding hierarchical local structures and spatially-local correlations.

Generally, the above 3D object recognition methods \cite{Qi_2017_CVPR, NIPS2017_7095, yan2020pointasnl} are modeled in a static impractical scenario, in which the categories of 3D objects are fixed and cannot change over time. However, in the real-world applications where the novel categories of 3D objects are often collected consecutively, these methods without substantial model modifications cannot learn new 3D objects in a streaming manner, thus causing significant performance degradation on old learned 3D objects (\emph{i.e.}, catastrophic forgetting \cite{10.1007/978-3-319-46493-0_37, Rebuffi_2017_CVPR, dong2022federated, wei2020lifelong}). To address it, a naive solution is to store all the training data of old and new categories, and retrain the current model again with a long delay. Nevertheless, this solution consumes high computational overhead and large memory occupancy, which is unacceptable in practical applications. 
For instance, the autonomous driving model \cite{Behl_2017_ICCV} trained with some specific recognition scenes cannot perform well in a prior-unknown environment where new 3D objects arrive continuously, due to the lack of continuous learning capacity for new 3D objects. Moreover, some visual navigation systems and intelligent robots \cite{8593741} working under a fixed application scenario may also suffer from catastrophic forgetting on previous learning experience, when collecting new 3D objects consecutively. 
The high computational complexity and large memory consumption make the autonomous driving model \cite{Behl_2017_ICCV}, visual navigation systems and intelligent robots \cite{8593741} impractical to access all 3D objects of old and new categories, thus resulting in forgetting on old 3D categories.

When tackling the above challenges, another feasible way is to attach a 3D point cloud feature extractor (\emph{e.g.}, PointNet \cite{Qi_2017_CVPR}, PointNet++ \cite{NIPS2017_7095}) to existing 2D class-incremental learning models \cite{Rebuffi_2017_CVPR, Castro_2018_ECCV, Wu_2019_CVPR, 8674766}. Unfortunately, this solution cannot effectively distinguish which 3D geometric characteristics within each class are essential to overcome the forgetting on old categories of 3D objects, due to the irregularity and deformable permutations of point cloud. The trained models fail to accurately recognize these old classes (\emph{i.e.}, catastrophic forgetting \cite{Castro_2018_ECCV, dong2022federated}) as new 3D categories arrive consecutively, since they capture the noisy characteristics or the common geometric properties among old 3D classes with similar structures. For instance, the above methods with PointNet \cite{Qi_2017_CVPR} or PointNet++ \cite{NIPS2017_7095} as feature extractor may highlight common structural layout among tables and chairs, but neglect the discriminative local geometric correlations within each class and further forget their characterizations under the class-incremental learning settings. Consequently, it is a challenging object recognition task to learn new 3D classes consecutively while alleviating forgetting on old classes.

To address the aforementioned issues, we design a novel \underline{In}cremental 3D \underline{O}bject \underline{R}ecognition \underline{Net}work (\emph{i.e.}, InOR-Net) for point cloud representation. The proposed InOR-Net model could not only learn new classes of 3D objects under a streaming manner, but also prevent the catastrophic forgetting on old 3D classes without using large infrastructures to store their training data. To be specific, we develop a category-guided geometric reasoning strategy, which captures the distinctive 3D characteristics within each class via constructing local geometric structures with category information as reference. Moreover, a novel critic-induced geometric attention mechanism is developed to identify which 3D geometric attributes within each class are important to prevent forgetting on old classes of 3D objects. It consists of a geometric attention network to quantify the contributions of 3D geometric characteristics, and a critical supervision network to feedback the quality of quantified contributions. To alleviate the catastrophic forgetting caused by class imbalance, we propose a dual adaptive fairness compensations strategy, which could correct the biased weights of the classifier via weight fairness compensation, and balance the biased predictions via score fairness compensation.

The proposed InOR-Net model achieves superior performance on three point cloud datasets, when compared with other baseline methods. In summary, we present the major contributions of this paper as follows:

\begin{itemize}
	\vspace{-3pt}
	\setlength{\itemsep}{0pt}		
	\setlength{\parsep}{1pt}	
	\setlength{\parskip}{0pt}
	
	\item We propose a novel Incremental 3D Object Recognition Network (\emph{i.e.}, InOR-Net) to recognize new categories of 3D objects continuously, without forgetting on old 3D categories. Our model is the first attempt to address 3D class-incremental learning in 3D object recognition field. 
	
	\item A category-guided geometric reasoning is designed to learn distinctive 3D characteristics within each class, which can be attained via exploring local geometric structures with category information as guidance.
	
	\item We develop a novel critic-induced geometric attention mechanism to distinguish which 3D geometric characteristics within each class are important to overcome the catastrophic forgetting on old classes, while neglecting the negative influence of useless 3D characteristics. 
	
	\item A dual adaptive fairness compensations mechanism is designed to address the forgetting brought by class imbalance. It corrects the biased weights of the classifier via weight fairness compensation, and balances the biased predictions via score fairness compensation. 
\end{itemize}

This work is a significant extension of our previous conference paper \cite{I3DOL_Dong_AAAI2021}. Compared with \cite{I3DOL_Dong_AAAI2021}, some main improvements of this paper could be presented below: 
\begin{enumerate}
\item  We ameliorate the construction manner of local geometric structures. We develop a category-guided geometric reasoning to reason distinctive 3D characteristics within each class by leveraging category information over all point cloud data rather than only an object used in \cite{I3DOL_Dong_AAAI2021}. 
\item We propose a novel critic-induced geometric attention to identify which 3D geometric characteristics within each class are useful to overcome the catastrophic forgetting on old classes of 3D objects. It could not only quantify their contributions, but also evaluate the quality of contributions to feedback the positive supervision. 
\item A dual adaptive fairness compensations strategy is designed to correct the biased weights of the classifier via weight fairness compensation in the training phase, while balancing the biased predictions on new classes of 3D objects via score fairness compensation in the test phase.
\item More comparison experiments on representative datasets illustrate the superior recognition performance of our InOR-Net model against other baseline frameworks. 
\end{enumerate}

\section{Related Work}\label{sec:related_work}

\subsection{3D Object Recognition}
There are many shape descriptors \cite{yan2020pointasnl, NIPS2017_7095} to characterize point cloud representation, but they are usually invariant to different transformations, thus the recognition task on different 3D classes cannot perform well. When Qi \emph{et al.} propose to apply convolutional neural network (\emph{i.e.}, PointNet \cite{Qi_2017_CVPR}) on point cloud data, deep learning based methods have achieved remarkable successes in 3D object recognition. For instance, PointNet++ \cite{NIPS2017_7095}, an extension version of PointNet \cite{Qi_2017_CVPR}, could encode hierarchical local structures of point clouds with increasing contextual scales. In addition, some permutation-robust networks \cite{Li_2018_CVPR, NIPS2018_7362} are proposed to explore the spatially-local relations in the orderless point cloud data. Wu \emph{et al.} \cite{Li_2018_CVPR} develop a point-to-node neighbor search mechanism to encode point cloud. Yan \emph{et al.} \cite{yan2020pointasnl} propose to eliminate the effect of outliers via an adaptive sampling. Unfortunately, the above methods suffer from large performance degradation in the real-world where the data of new 3D categories are collected consecutively.

\subsection{Class-Incremental Learning}
According to the methods with access to nothing, generated data or original data (\emph{i.e.}, exemplars) from the old classes, we divide the extensive studies about class-incremental learning \cite{10.1007/978-3-319-46493-0_37, Oleksiy_2019_CVPR, Wu_2019_CVPR, wei2021incremental222} on 2D image classification \cite{chen2019two} into three categories. For example, when the training data of old classes is unavailable,  \cite{10.1007/978-3-319-46493-0_37, Shmelkov_2017_ICCV} propose the distillation strategy \cite{yu2022cyclic} to overcome the catastrophic forgetting on old classes. 
Yang \emph{et al.} \cite{8674766} propose a consecutive updating mechanism for concept drift detection. Leo \emph{et al.} \cite{9459438} develop a classification confidence threshold to selectively finetune neural network for anti-forgetting. Wei \emph{et al.} \cite{9877899} focus on exploring class-disentangled representations to tackle forgetting on old classes. Yu \emph{et al.} \cite{9737321} introduce a self-training strategy to perform continual semantic segmentation tasks with limited forgetting. \cite{9533187} designs one unified framework to address continual new classes with efficient training of model parameters. 
Kirkpatrick \emph{et al.} \cite{Kirkpatrick3521} introduce new regulators to restrict the model's optimization caused by new classes, which maintains its memory for old classes. In addition, different kinds of generative adversarial networks \cite{9616392_Dong} are utilized by \cite{Oleksiy_2019_CVPR, NIPS2018_7836} to produce synthetic samples for old categories.

For the exemplar replay, there are usually a small quantity of exemplars available for each old category \cite{Rebuffi_2017_CVPR}. In this case, class imbalance becomes a serious challenge \cite{Castro_2018_ECCV, deesil-eccv2018}. Belouadah \emph{et al.} \cite{Belouadah_2019_ICCV} provide a memory with negligible storage cost to store statistical information of old categories. With the training data of new categories arriving, \cite{DBLP:journals/corr/RusuRDSKKPH16} proposes to expand the neural network in a progressive manner. To address the large-scale class-incremental learning, Wu \emph{et al.} \cite{Wu_2019_CVPR} focus on correcting the deviation of new categories within the classifier network. A random path selection strategy is proposed by Rajasegaran \emph{et al.} \cite{NIPS2019_9429} to choose optimal information flow for new tasks. Simon \emph{et al.}  \cite{Christian2021MGeoCont} ameliorate the knowledge distillation technology, and measure the similarity between previous and current responses via geodesic path. Hu \emph{et al.} \cite{Hu_20121_CVPR} aim to explore the causal effect between the old and new categories. However, due to the noisy permutations, missing structures and irregularity of point cloud, these methods \cite{9877899, 9533187, 9737321, Christian2021MGeoCont, Hu_20121_CVPR} cannot effectively distinguish which 3D geometric characteristics within each class are beneficial to prevent catastrophic forgetting on old class of 3D objects.

\section{The Proposed Model}
\subsection{Problem Definition and Overview}

\textbf{Problem Definition:}
For the experimental configurations about 3D class-incremental learning, we follow the standard settings widely-used in 2D field \cite{Rebuffi_2017_CVPR, Castro_2018_ECCV, Wu_2019_CVPR, NIPS2019_9429, Christian2021MGeoCont, Hu_20121_CVPR}. 
Specifically, we define the streaming training data including total $S$ incremental states as $D = \{D_1, D_2, \cdots, D_S\}$. In the $s$-th $(s=1, 2, \cdots, S)$ incremental state, the training subset $D_s = \{x_i^s, y_i^s\}_{i=1}^{n_s}$ consists of $n_s$ point clouds, where $x_i^s\in\mathbb{R}^{U\times 3}$ is the $i$-th point cloud including $U$ sampled three-dimension points, and $y_i^s$ represents its one-hot category label. The labels $\{y_i^s\}_{i=1}^{n_s}$ include $K_s$ new classes in the $s$-th incremental state. They cannot overlap with $K_p = \sum_{i=1}^{s-1}K_i$ old classes that are learned in previous $s-1$ incremental states. The training subset $\{D_i\}_{i=1}^S$ with different new classes arrives consecutively in this paper. 
Similar to \cite{Rebuffi_2017_CVPR, Castro_2018_ECCV, Wu_2019_CVPR, NIPS2019_9429, Christian2021MGeoCont, Hu_20121_CVPR}, we select a small quantity of samples from $K_p$ old classes of 3D objects to construct the exemplar set $M$. With access to the subset $D_s$ and the selected exemplar set $M$, the prediction task is to classify both $K_p$ old categories and $K_s$ new categories in the $s$-th state. The number of exemplars from $M$ is small (\emph{i.e.}, $|M|/K_p\ll n_s/K_s$) in our experiments.

\begin{figure*}[t]
	\centering
	\includegraphics[width=515pt, height =190pt]
	{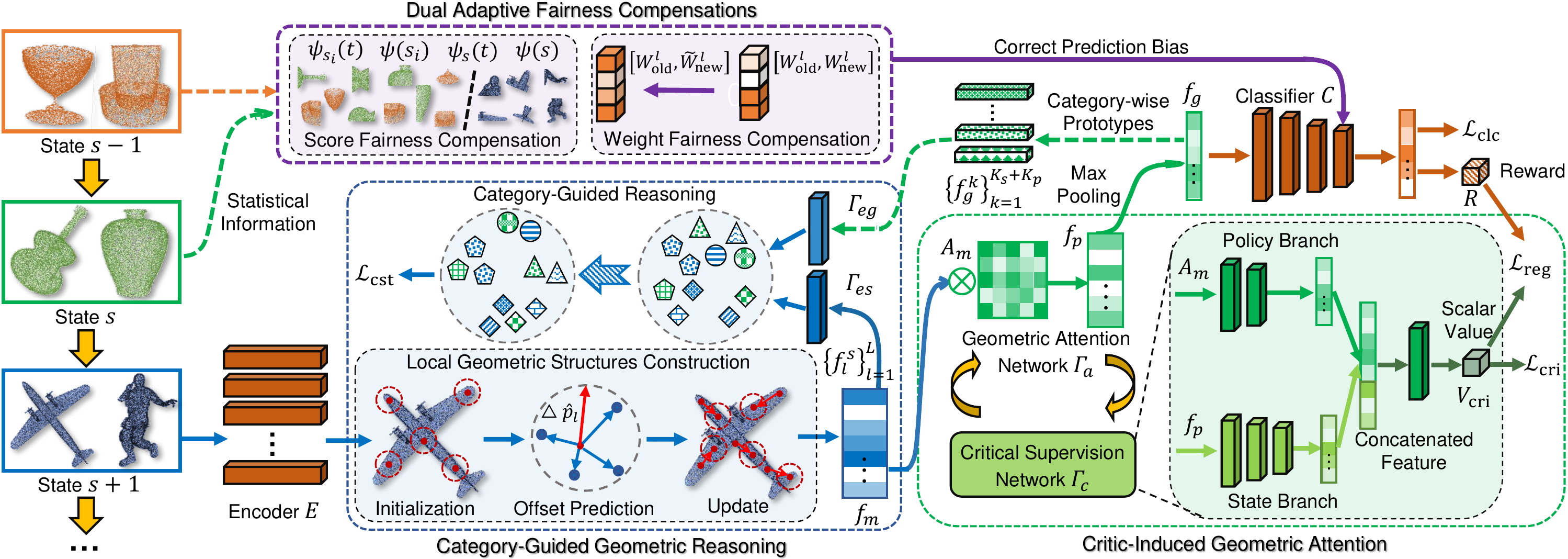}
	\vspace{-17pt}
	\caption{Graphical pipeline of our InOR-Net model, including the \textit{category-guided geometric reasoning} (the blue block) to capture distinctive 3D geometric characterizations within each class, the \textit{critic-induced geometric attention} (the green block) to identify which 3D geometric characteristics are beneficial to overcome catastrophic forgetting on old classes of 3D objects, and the \textit{dual adaptive fairness compensations} (the purple block) to address the forgetting brought by class imbalance between old and new categories. } 
	\label{fig:overview_of_our_model}
\end{figure*}

\textbf{Overview:}
The graphical pipeline of our InOR-Net model is depicted in Fig.~\ref{fig:overview_of_our_model}. First, we forward the 3D object from $D_s\cup M$ into the encoder $E(\cdot)$ to extract its low-level feature. Subsequently, the extracted feature is fed into the category-guided geometric reasoning to capture distinctive 3D geometric characterizations within each class under the guidance of category information, and then we obtain its mid-level feature $f_m$ via optimizing $\mathcal{L}_{\mr{cst}}$. Afterwards, $f_m$ is forwarded into the critic-induced geometric attention to evaluate which 3D geometric characteristics are beneficial to overcome catastrophic forgetting on old classes of 3D objects, while filtering out the useless geometric information via optimizing $\mathcal{L}_{\mr{cri}}$ and $\mathcal{L}_{\mr{reg}}$. We apply max-pooling function to $f_p$ with geometric attention and get a global feature $f_g$, which is then passed into the classifier $C(\cdot)$ to make predictions via minimizing $\mathcal{L}_{\mr{clc}}$.  
Moreover, the dual adaptive fairness compensations could further compensate biased weights of the classifier and biased score predictions that are brought by class imbalance between old and new categories. Our InOR-Net model could alleviate the forgetting on old classes via remembering their distinctive 3D geometric characterizations.

\subsection{Category-Guided Geometric Reasoning}\label{sec:category_guided_geometric_reasoning}
Generally, each point cloud can be characterized by $L$ local geometric structures, which correspond to $L$ point subsets $\big\{P_l|P_l = \{\hat{p}_l, p_{l1}, p_{l2}, \cdots, p_{lm}\in\mathbb{R}^3\}\big\}_{l=1}^L$ in the point cloud. $\hat{p}_l$ is the centroid of the $l$-th local geometric structure $P_l$, and $\{p_{l1}, p_{l2}, \cdots, p_{lm}\}$ are $m$ nearest neighbor points surrounding around $\hat{p}_l$. Obviously, $\hat{p}_l$ determines the location of the $l$-th local geometric structure $P_l$ and its $m$ nearest neighbor points.

To capture local context information, most existing researches \cite{NIPS2017_7095, NIPS2018_7362} utilize the random sampling or farthest point sampling strategy to select the centroids for local geometric structures $\{P_l\}_{l=1}^L$. Though these researches guarantee that the selected centroids fully consider the entire point cloud, they cannot cover the local structures with distinctive 3D characteristics (\emph{e.g.}, the tail in airplane, the geometric layout in chair, etc.) that are more effective to characterize the point cloud. Moreover, they aim to explore distinctive 3D characteristics from an object, but cannot make sure that the local structures within each object can capture the discriminative category information. For example, there are many types of chairs in the real-world, which not only share the common structure layout among all kinds of chairs, but also have unique properties for each kind of chairs. However, the existing works \cite{NIPS2017_7095, NIPS2018_7362} focus on exploring the unique properties within each chair, while neglecting the discriminative category information (\emph{e.g.}, common structure layout) to distinguish the chair and other similar categories (\emph{e.g.}, desk, table, etc). Due to the above limitations, the performance of \cite{NIPS2017_7095, NIPS2018_7362} severely suffers from the noisy points (\emph{i.e.}, noisy local structures).

To address these limitations, as shown in Fig.~\ref{fig:overview_of_our_model}, we develop the category-guided geometric reasoning, which captures local structures with distinctive 3D characteristics by considering category information over all point cloud data rather than only an object. Intuitively, for 3D class-incremental learning, these distinctive 3D geometric characteristics within each class are essential to alleviate catastrophic forgetting on old classes. To this end, we first present how to adaptively construct $L$ local geometric structures $\{P_l\}_{l=1}^L$, and then consider semantic category information as guidance (\emph{i.e.}, category-guided reasoning) to capture distinctive 3D geometric characteristics.

$\bullet$ \textbf{Local Geometric Structures Construction:} 
\cite{NIPS2017_7095, NIPS2018_7362} initialize the locations of $L$ local geometric structures via farthest point sampling, and utilize the fixed centroids of local geometric structures to encode semantic context. Different from them, we adaptively modify the centroids of local structures (\emph{i.e.}, the geometric offset prediction) as the training process. Compared with deformable convolution \cite{DBLP:journals/corr/DaiQXLZHW17} using semantic features of 2D images for offset prediction, we employ the edge vectors of each local geometric structure as reference to learn geometric offset of the centroid via network itself \cite{I3DOL_Dong_AAAI2021, NIPS2019_8940}. To be specific, we first quantify the semantic context of each edge as the contribution weight, and then integrate the weighted edge vectors of local geometric structures together to achieve offset prediction of the centroid. In other words, for each local geometric structure, the voting strategy of surrounding edge vectors with different contributions determines the centroid offset. Thus, given the $l$-th local geometric structure $P_l$ constructed via farthest point sampling \cite{NIPS2017_7095, NIPS2018_7362}, the geometric offset $\triangle\hat{p}_l$ of $\hat{p}_l$ is formulated as: 
\begin{align}
	\triangle \hat{p}_l = \frac{1}{m} \sum\limits_{i=1}^m \big(\Gamma_o((\hat{f}_l-f_{li}); \theta_{\Gamma_o}) \cdot (\hat{p}_l-p_{li}) \big),
	\label{eq:offset_prediction}
\end{align}
where $\Gamma_o(\cdot)$ transforms the semantic knowledge of edge vectors to scale contribution weights via a convolution layer, and $\theta_{\Gamma_o}$ denotes its parameters. $(\hat{p}_l-p_{li})$ represents the $i$-th local edge vector. $\hat{f}_l\in\mathbb{R}^{d_p}$ and $\{f_{li}\in\mathbb{R}^{d_p}\}_{i=1}^m$ are the semantic features of $\hat{p}_l$ and its $m$ nearest neighbors $\{p_{li}\}_{i=1}^m$, respectively. 
$d_p$ is the feature dimension of $\{f_{li}\}_{i=1}^m$ that are extracted via the encoder $E(\cdot)$, as shown in Fig.~\ref{fig:overview_of_our_model}.

To update the $l$-th local geometric structure, we add the offset vector $\triangle\hat{p}_l$ in Eq.~\eqref{eq:offset_prediction} to the original centroid $\hat{p}_l$, and reselect $m$ nearest neighbors $\{p_{l1}, p_{l2}, \cdots, p_{lm}\}$ surrounding around the new centroid $\hat{p}_l$:
\begin{align}
&\hat{p}_l = \hat{p}_l + \triangle\hat{p}_l, \nonumber \\
&\{p_{l1}, p_{l2}, \cdots, p_{lm}\} = \mr{knn}(\hat{p}_l|p_j\in\mathbb{R}^3, j=1,\cdots, U),
\label{eq:local_structure_update}
\end{align}
where $\mr{knn}(\cdot)$ \cite{1053964, 10.1007/978-3-540-74976-9_25} is employed to search $m$ nearest neighbors around the new $\hat{p}_l$ by traversing the whole point cloud $\{p_j\}_{j=1}^U$. 
$U$ is the number of points sampled from a point cloud, and we follow \cite{Qi_2017_CVPR} to set $U=1024$ in this paper.
The semantic representation $f_l^s\in\mathbb{R}^{d_s}$ of the $l$-th local geometric structure can then be determined by encoding the representations of $m$ nearest neighbor points:
\begin{align}
f_l^s = \max_{i=1,2,\cdots,m} \Gamma_s(f_{li}; \theta_{\Gamma_s}),
\label{eq:update_structure_feature}
\end{align}
where $\{f_{li}\}_{i=1}^m$ denote the representations of $m$ nearest neighbor points around the updated $\hat{p}_l$. $d_s$ is the feature dimension of $f_l^s\in\mathbb{R}^{d_s}$ that is encoded via $\Gamma_s(\cdot)$. 
$\Gamma_s(\cdot)$ is a convolutional layer encoding the representations of all nearest neighbor points, and its parameters are denoted as $\theta_{\Gamma_s}$. Thus, the features $\{f_{l}^s\in\mathbb{R}^{d_s}\}_{l=1}^L$ of $L$ local geometric structures in each point cloud can be obtained via Eq.~\eqref{eq:update_structure_feature}. Inspired by \cite{NIPS2017_7095}, we concatenate $\{f_l^s\}_{l=1}^L$ of local geometric structures as  $f_m\in\mathbb{R}^{L\times d_s}$, and regard it as the low-level representations over the whole point cloud.

$\bullet$ \textbf{Category-Guided Reasoning:}
To further help the local geometric structures capture distinctive 3D characteristics within each class, we employ the global category-wise prototypes as guidance, and perform a self-supervised semantic consistency between local structures and category-wise prototypes for category-guided reasoning. To this end, as shown in Fig.~\ref{fig:overview_of_our_model}, we perform the max-pooling operation on low-level representations with geometric attention (\emph{i.e.}, $f_p$), and obtain its global feature $f_g\in\mathbb{R}^{d_s}$ for the whole point cloud. Given a mini-batch $T_B = \{x_i^s, y_i^s\}_{i=1}^B (B \ll n_s)$ with $B$ point clouds sampled from the $s$-th incremental state, the estimated category-wise prototype $\hat{f}_g^k$ of the $k$-th category is formulated as the mean feature of all global representations belonging to the $k$-th category ($k=1, 2, \cdots, K_s+K_p$) in $T_B$: 
\begin{align}
\hat{f}_g^k = \mathbb{E}_{(x_i^s, y_i^s)\in T_B} [\frac{1}{N_t}\sum_{i=1}^B f_{gi} \cdot \mathbf{1}_{\arg\max y_i^s = k}],
\label{eq:class_prototype}
\end{align}
where $N_t = \sum_{i=1}^B \mathbf{1}_{\arg\max y_i^s = k}$ denotes the number of point clouds belonging to the $k$-th class. $f_{gi}\in\mathbb{R}^{d_s}$ is the $i$-th point cloud's global representation. To eliminate the sampling randomness of mini-batch, we construct global category-wise prototype $f_g^k$ for the $k$-th class via an exponential update:
\begin{align}
f_g^k = \gamma f_g^k + (1-\gamma) \hat{f}_g^k, 
\label{eq:global_prototype}
\end{align}
where $\gamma=0.7$ is the balanced weight.

The local representations $\{f_{l}^s\in\mathbb{R}^{d_s}\}_{l=1}^L$ and global category-wise prototypes $\{f_{g}^k\in\mathbb{R}^{d_s}\}_{k=1}^{K_s+K_p}$ may have some semantic heterogeneity \cite{yang2020heterogeneous, zhang2021practical, shen2020federated}, so we employ convolutional networks $\Gamma_{es}(\cdot)$ and $\Gamma_{eg}(\cdot)$ to embed them into a semantic-shared feature space. When the input point cloud belongs to the $k$-th class, we enforce the embedded representations $\{\Gamma_{es}(f_{l}^s; \theta_{\Gamma_{es}})\in\mathbb{R}^{d_c}\}_{l=1}^L$ of local geometric structures to be closer to the embedded global category-wise prototype $\Gamma_{eg}(f_{g}^k; \theta_{\Gamma_{eg}})\in\mathbb{R}^{d_c}$ of the $k$-th class, while maximizing their dissimilarity with other embedded global prototypes $\{\Gamma_{eg}(f_{g}^i, \theta_{\Gamma_{eg}})\in\mathbb{R}^{d_c}| i\neq k\}_{i=1}^{K_s+K_p}$ via optimizing $\mathcal{L}_{\mr{cst}}$:
\begin{align}
&\mathcal{L}_{\mr{cst}}^l = \log\big(1+ \sum\limits_{i\neq k} \exp\big(\tau \mathcal{N}(\Gamma_{es}(f_{l}^s; \theta_{\Gamma_{es}}))^{\top} \mathcal{N}(\Gamma_{eg}(f_{g}^i; \theta_{\Gamma_{eg}})) \nonumber \\
& \qquad\qquad\qquad\quad - \tau\mathcal{N}(\Gamma_{es}(f_{l}^s; \theta_{\Gamma_{es}}))^{\top} \mathcal{N}(\Gamma_{eg}(f_{g}^k; \theta_{\Gamma_{eg}})) \big)\big), \nonumber \\
& \mathcal{L}_{\mr{cst}} = \mathbb{E}_{(x_i^s, y_i^s)\in D_s\cup M}\sum\nolimits_{l=1}^L \mathcal{L}_{\mr{cst}}^l,
\label{eq:semantic_consistency}
\end{align}
where $\theta_{\Gamma_{eg}}$ and $\theta_{\Gamma_{es}}$ denote the network parameters. $d_c$ is the feature dimension of embedding space. $\mathcal{N}(x) = \frac{x - \mr{mean}(x)}{\|x\|}$ represents a $\ell_2$ normalization function, and $\mr{mean}(x)$ denotes the mean value of $x$.

Before performing a self-supervised semantic consistency metric via Eq.~\eqref{eq:semantic_consistency}, we first normalize the embedded representations $\Gamma_{es}(f_{l}^s; \theta_{\Gamma_{es}})$ and $\Gamma_{eg}(f_{g}^k; \theta_{\Gamma_{eg}})$ via $\ell_2$ normalization function $\mathcal{N}(\cdot)$, and then utilize a scale value $\tau=64$ to re-scale the embedded representations.  
Existing works \cite{9449988, Liu_2017_CVPR} have shown the re-scaling strategy could stabilize the training process and make 3D characteristics within each class more discriminative. Intuitively, when minimizing Eq.~\eqref{eq:semantic_consistency}, we enforce local geometric structures to explore unique 3D characteristics within each class, while neglecting the properties common with other classes. These distinctive 3D characteristics are essential to alleviate forgetting on old classes of 3D objects.

\subsection{Critic-Induced Geometric Attention}\label{sec:critical_geometric_aware_attention}
Although the category-guided geometric reasoning could construct $L$ discriminative local geometric structures to capture distinctive 3D characteristics, these characteristics contributes unequally to address the catastrophic forgetting on old classes. That is to say, some local geometric structures containing more common 3D characteristics may strengthen the forgetting, while the others with more distinctive 3D characteristics would alleviate the forgetting. To address this issue, as presented in Fig.~\ref{fig:overview_of_our_model}, we develop the critic-induced geometric attention to emphasize unique 3D geometric characteristics within each class, while mitigating the forgetting caused by common 3D properties. Specifically, we first use the geometric attention network $\Gamma_a(\cdot)$ to calibrate the contributions of local geometric structures $\{P_l\}_{l=1}^L$, and further introduce the critical supervision network $\Gamma_c(\cdot)$ as guidance to evaluate the quality of the contributions quantified via $\Gamma_a(\cdot)$. $\Gamma_c(\cdot)$ can provide positive supervision to guide $\Gamma_a(\cdot)$ maximize the contribution gain, even though misleading quantification of $\Gamma_a(\cdot)$ appears.

$\bullet$ \textbf{Geometric Attention Network $\Gamma_a(\cdot)$:}
According to the vanilla attention strategy \cite{fu2019dual} in image understanding, we introduce a residual learning to the geometric attention network $\Gamma_a(\cdot)$. It can calibrate the weights of different geometric structures. Therefore, we obtain the final low-level semantic representation $f_p\in\mathbb{R}^{L\times d_s}$ via the attention mechanism: 
\begin{align}
	f_p & = \mathcal{A}_m\odot f_m + f_m \nonumber \\
	 & = \psi_s\big(\Gamma_u(\psi_r(\Gamma_d(f_m; \theta_{\Gamma_d})); \theta_{\Gamma_u})\big)\odot f_m + f_m,
	\label{eq:centroid_attention}
\end{align}
where
$\mathcal{A}_m = \psi_s\big(\Gamma_u(\psi_r(\Gamma_d(f_m; \theta_{\Gamma_d})); \theta_{\Gamma_u})\big)\in\mathbb{R}^{L\times d_s}$ represents the geometric attention (\emph{a.k.a.} the contributions of local geometric structures). 
$\odot$ is the Hadamard product. $\psi_s(\cdot)$ and $\psi_r(\cdot)$ are the sigmoid and ReLU functions. $\Gamma_u(\cdot)$ and $\Gamma_d(\cdot)$ denote the channel-upscaling and channel-downscaling layers with the ratio as $r=4$. $\theta_{\Gamma_u}$ and $\theta_{\Gamma_d}$ are their weights, which are both denoted as the parameters $\theta_{\Gamma_a}$ of $\Gamma_a(\cdot)$ for simplification. As shown in Fig.~\ref{fig:overview_of_our_model}, we then perform max-pooling on $f_p$ to extract global feature $f_g\in\mathbb{R}^{d_s}$, and forward it into $C(\cdot)$ for object recognition.

$\bullet$ \textbf{Critical Supervision Network $\Gamma_c(\cdot)$:}
As aforementioned, $\Gamma_c(\cdot)$ focuses on maximizing the gain of geometric attention $\mathcal{A}_m$ over the basic network via a task reward strategy. To this end, as depicted in Fig.~\ref{fig:overview_of_our_model}, the critical supervision network $\Gamma_c(\cdot)$ takes both $f_p$ and $\mathcal{A}_m$ as the inputs. It contains two branches, \emph{i.e.,} a state branch with a convolutional block, a flatten operation and two fully-connected layers to extract semantic context from $f_p$; and a policy branch with a convolutional block, a flatten operation and a fully-connected layer to encode quantified contribution $\mathcal{A}_m$. Afterwards, the outputs of both state and policy branches are concatenated together, and are fed into a fully-connected layer to obtain a scalar gain $V_{\mr{cri}} = \Gamma_c(f_p, \mathcal{A}_m; \theta_{\Gamma_c})$, where $\theta_{\Gamma_c}$ is the network weights. Two losses (\emph{i.e.}, the critic loss $\mathcal{L}_{\mr{cri}}$ and regression loss $\mathcal{L}_{\mr{reg}}$) are designed to maximize the gain with positive guidance.

\textbf{1. Critic Loss $\mathcal{L}_{\mr{cri}}$:}
For training the geometric attention network $\Gamma_a(\cdot)$, the critic loss $\mathcal{L}_{\mr{cri}}$ is proposed to maximize the scalar gain value $V_{\mr{cri}}$ with respect to the geometric attention $\mathcal{A}_m$ (\emph{i.e.}, minimizing $\mathcal{L}_{\mr{cri}}$ in Eq.~\eqref{eq:critic_loss}):
\begin{align}
\mathcal{L}_{\mr{cri}} = \mathbb{E}_{f_p\in D_s\cup M}[-V_{\mr{cri}}] 
= \mathbb{E}_{f_p\in D_s\cup M}[-\Gamma_c(f_p, \mathcal{A}_m; \theta_{\Gamma_c})]. 
\label{eq:critic_loss} 
\end{align}
Intuitively, $\mathcal{L}_{\mr{cri}}$ encourages $\Gamma_c(\cdot)$ to highlight the local geometric structures with higher contribution scores. It maximizes the positive gain by generating higher gain value $V_{\mr{cri}}$.

\textbf{2. Regression Loss $\mathcal{L}_{\mr{reg}}$:}
The regression loss $\mathcal{L}_{\mr{reg}}$ is designed to guide $\Gamma_c(\cdot)$ feedback accurate supervision for $\Gamma_a(\cdot)$ via a new reward $R$. Specifically, $R$ is composed of a classification reward $R_c$ and an amelioration reward $R_a$, \emph{i.e.}, $R=R_c + R_a$. $R_c$ measures whether the geometric attention $\mathcal{A}_m$ learned via $\Gamma_a(\cdot)$ leads to correct prediction, which is formulated as follows: 
\begin{equation}
\begin{split}
R_c =
\left\{
\begin{aligned}		
&1, \quad\quad \mr{if}~\arg\max C(f_g; \theta_C)  = \arg\max y, \\
&0, \qquad\qquad\qquad\quad \mr{otherwise}, \\
\end{aligned} 					
\right.											
\end{split}							
\label{equation:definition_R^s}	
\end{equation}
where $\theta_{C}$ denotes the parameters of classifier $C(\cdot)$, and $C(f_g; \theta_C)$ is the probability outputs of global feature $f_g$ with the geometric attention $\mathcal{A}_m$. $y\in D_s\cup M$ is the one-hot groundtruth of the input point cloud. Moreover, the amelioration reward $R_a$ examines whether $\mathcal{A}_m$ learned via $\Gamma_a(\cdot)$ facilitates the positive prediction, where $R_a$ is defined as:
\begin{align}
R_a =
\left\{
\begin{aligned}		
&1,\;  \mr{if}~C(f_g; \theta_C)^k \textgreater C(f_{g'}; \theta_C)^k, k = \arg\max y   \\
&0, \;\;\qquad\qquad\qquad \mr{otherwise},  \\
\end{aligned} 					
\right.																
\label{equation:definition_R^a}	
\end{align}
where $C(f_g; \theta_C)^k$ and $C(f_{g'}; \theta_C)^k$ respectively represent the $k$-th category probability predicted by the classifier $C(\cdot)$ with geometric attention $\mathcal{A}_m$ or not. Here, $f_{g'}$ can be obtained when we directly perform max-pooling operation on $f_m$ without using $\mathcal{A}_m$. We then develop a reward regression loss $\mathcal{L}_{\mr{reg}}$ to guide $\Gamma_c(\cdot)$ feedback accurate supervision. $\mathcal{L}_{\mr{reg}}$ aims to minimize the gap between the estimated scalar gain $V_{\mr{cri}}$ and defined reward $R$:
\begin{equation}
\mathcal{L}_{\mr{reg}} = \mathbb{E}_{f_p\in D_s\cup M}\big[(V_{\mr{cri}} - R)^2],
\label{eq:reward_regression_loss}
\end{equation}
where $V_{\mr{cri}} = \Gamma_c(f_p, \mathcal{A}_m; \theta_{\Gamma_c})$, and $R=R_c + R_a$.

\subsection{Dual Adaptive Fairness Compensations}\label{sec:dual_adaptive_fairness_compensations}
Although the above modules can explore distinctive 3D characteristics within local geometric structures for each class, the classifier $C(\cdot)$ is easily prone to forget the old classes since there is a severe class imbalance issue between old and new categories of 3D objects ($|M|\ll n_s$). To address this issue, most 2D researches \cite{Castro_2018_ECCV, Rebuffi_2017_CVPR, Wu_2019_CVPR, NIPS2019_9429} propose to pay more attention on old classes via the knowledge distillation strategy. However, they cannot prevent the fully-connected layers of classifier $C(\cdot)$ from being highly biased, thus strengthening the forgetting on old classes. The classifier $C(\cdot)$ often predicts well on new classes with a large number of 3D objects, while performing badly on old classes without abundant data.

To overcome above issues, the dual adaptive fairness compensations strategy is designed to correct prediction bias among the old and new categories of 3D objects. It consists of a weight fairness compensation to correct the biased weights of classifier $C(\cdot)$ in the training phase, and a score fairness compensation to balance the biased predictions on new 3D categories in the test phase.

\begin{algorithm}[t]
	\caption{Training Pipeline of Our InOR-Net Model. }
	\begin{algorithmic}[1]
		\State {\bfseries Input:} The subset $D_s = \{x_i^s, y_i^s\}_{i=1}^{n_s}$ including the data of new categories, the exemplar set $M$, and $\{\lambda_1, \lambda_2\}$.
		\State {\bfseries Initialize: $\{\theta_{E}, \theta_{C}, \theta_{\Gamma_c}, \theta_{\Gamma_a}\}$};
		\State {\bfseries While} not converged \textbf{do}
		\State \hspace{0.1cm} Construct a mini-batch $T_B = \{x_i^s, y_i^s\}_{i=1}^B$ from $D_s\cup M$;
		\State \hspace{0.1cm} Update $\{\theta_{E}, \theta_{C}\}$ via minimizing $\mathcal{L}_{\mr{clc}}+\lambda_2\mathcal{L}_{\mr{cst}}$;
		\State \hspace{0.1cm} Update $\theta_{\Gamma_a}$ via minimizing $\mathcal{L}_{\mr{clc}}+\lambda_1\mathcal{L}_{\mr{cri}}+\lambda_2\mathcal{L}_{\mr{cst}}$;		
		\State \hspace{0.1cm} Update $\theta_{\Gamma_c}$ via minimizing $\mathcal{L}_{\mr{reg}}$;
		\State {\bfseries End}
		\State Store statistical information to compensate the biased predictions via Eq.~\eqref{eq:score_fairness_compensation^a} in the inference phase;
		\State {\bfseries Return} $\{\theta_{E}, \theta_{C}, \theta_{\Gamma_c}, \theta_{\Gamma_a}\}$;
	\end{algorithmic}
	\label{alg:InORNet}	
\end{algorithm}

$\bullet$ \textbf{Weight Fairness Compensation:}
Denote the weight $W^l$ of the last fully-connected layer in the classifier $C(\cdot)$ as $W^l = [W_{\mr{old}}^l, W_{\mr{new}}^l]\in\mathbb{R}^{d_w\times(K_p+K_s)}$, where $K_p$ and $K_s$ respectively denote the numbers of old and new categories, and $d_w$ is the input feature dimension of last fully-connected layer in $C(\cdot)$. The weight matrices $W_{\mr{old}}^l$ and $W_{\mr{new}}^l$ are defined as $W_{\mr{old}}^l = [w_1, w_2, \cdots, w_{K_p}]\in\mathbb{R}^{d_w\times K_p}$ and $W_{\mr{new}}^l = [w_{K_p+1}, w_{K_p+2}, \cdots, w_{K_p+K_s}]\in\mathbb{R}^{d_w\times K_s}$. The corresponding norms of the weight matrices $W_{\mr{old}}^l$ and $W_{\mr{new}}^l$ are written as $N_{\mr{old}} = [||w_1||, ||w_2||, \cdots, ||w_{K_p}||]\in\mathbb{R}^{K_p}$ and $N_{\mr{new}} = [||w_{K_p+1}||, ||w_{K_p+2}||, \cdots, ||w_{K_p+K_s}||]\in\mathbb{R}^{K_s}$. Therefore, the normalized weight matrix $\tilde{W}_{\mr{new}}^l$ for new classes of 3D objects is formulated as $\tilde{W}_{\mr{new}}^l = \frac{\mr{mean}(N_{\mr{old}})}{\mr{mean}(N_{\mr{new}})}\cdot W_{\mr{new}}^l$. After applying weight fairness compensation, the corrected weight $\tilde{W}^l$ of the last fully-connected layer is:
\begin{align}
	\tilde{W}^l = [W_{\mr{old}}^l, \tilde{W}_{\mr{new}}^l] = [W_{\mr{old}}^l, \frac{\mr{mean}(N_{\mr{old}})}{\mr{mean}(N_{\mr{new}})}\cdot W_{\mr{new}}^l].
	\label{eq:weight_fairness_compensation}
\end{align}

\begin{table}[t]
	\centering
	\setlength{\tabcolsep}{1.3mm}
	\caption{Architecture details of our proposed InOR-Net model. } 
	\scalebox{0.8}{
		\begin{tabular}{|c|cc|cccc|}
			\hline
			Network & Layers & Transformation & Kernel & Input Channel & Output Channel & Stride \\
			\hline
			\multirow{3}{*}{Encoder $E(\cdot)$} & \multirow{3}{*}{3} & T-Net \cite{Qi_2017_CVPR} & 3 & 3 & 64 & 1 \\
			&  & T-Net \cite{Qi_2017_CVPR} & 1 & 64 & 128 & 1 \\
			&  & \ding{55} & 1 & 128 & 512 & 1 \\  
			\hline	
			$\Gamma_{o}(\cdot)$ & 1 & \ding{55} & 1 & 512 & 3 & 1 \\
			$\Gamma_{s}(\cdot)$ & 1 & \ding{55} & 1 & 512 & 1024 & 1 \\		 
			$\Gamma_{eg}(\cdot)$ & 1 & \ding{55} & 1 & 1024 & 256 & 1 \\
			$\Gamma_{es}(\cdot)$ & 1 & \ding{55} & 1 & 1024 & 256 & 1 \\
			$\Gamma_{u}(\cdot)$ & 1 & \ding{55} & 1 & 256 & 1024 & 1 \\
			$\Gamma_{d}(\cdot)$ & 1 & \ding{55} & 1 & 1024 & 256 & 1 \\
			\hline
			\multirow{2}{*}{$\Gamma_c(\cdot)$ (Policy)} & \multirow{2}{*}{2} & \ding{55} & 3 & 1024 & 256 & 1  \\
			&  & \ding{55} & 1 & 256$\cdot$L & 64 & 1 \\
			\hline
			\multirow{3}{*}{$\Gamma_c(\cdot)$ (State)} & \multirow{3}{*}{3} & \ding{55} & 3 & 1024 & 256 & 1  \\
			&  & \ding{55} & 1 & 256$\cdot$L & 256 & 1  \\ 
			&  & \ding{55} & 1 & 256 & 64 & 1  \\ 
			\hline
		\end{tabular}
	} 	
	\label{tab:network_implementation} 
\end{table}

\begin{figure}[t]
	\centering
	\includegraphics[width =250pt, height=152pt] {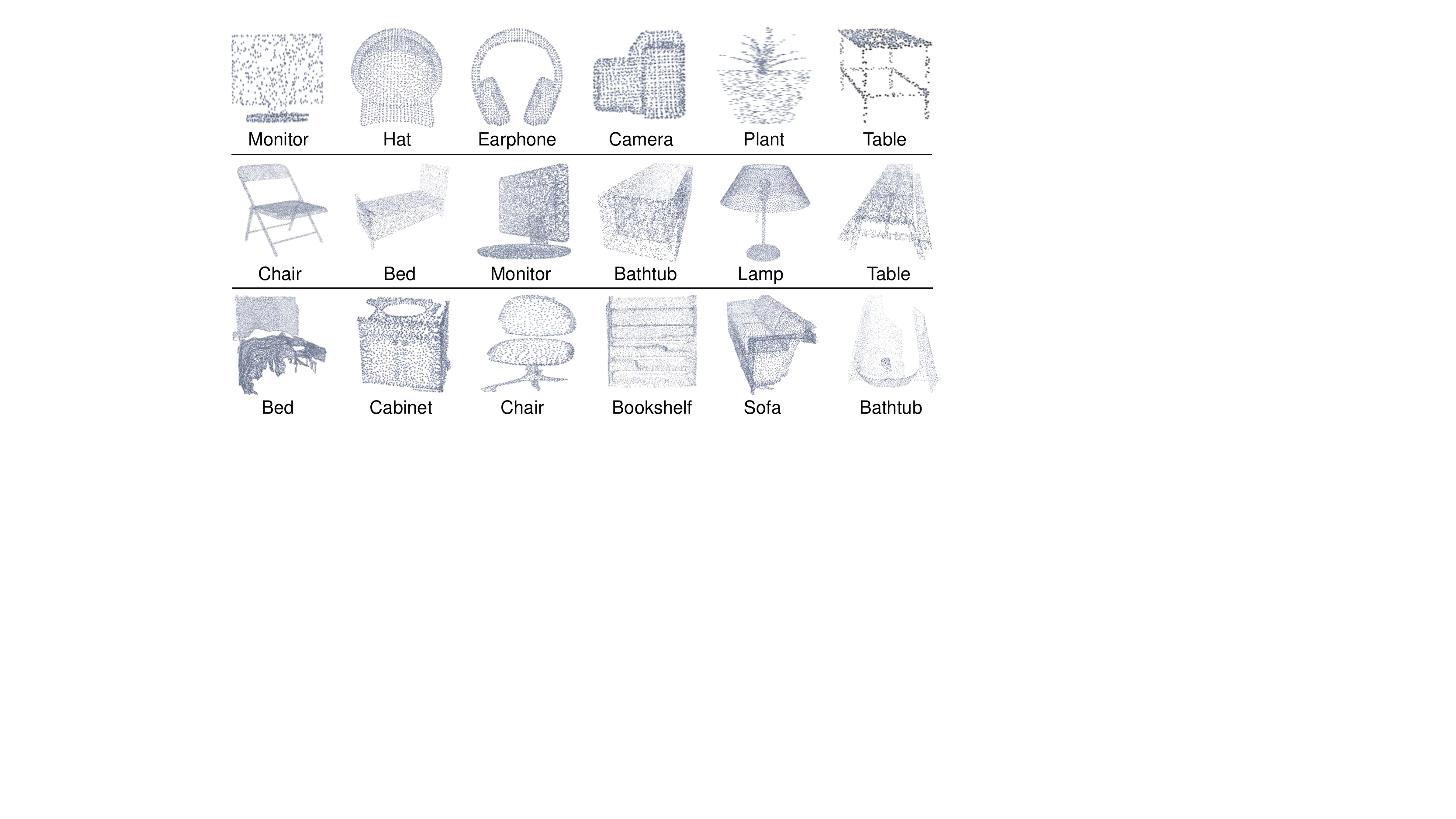}
	\caption{Visualization of some 3D objects in three benchmark datasets, where the top, middle and bottom rows respectively denote the samples collected from ShapeNet \cite{DBLP:journals/corr/ChangFGHHLSSSSX15}, ModelNet \cite{7298801} and ScanNet \cite{8099744}. }
	\label{fig:example_samples}
\end{figure}

\begin{table*}[t]
	\centering
	\setlength{\tabcolsep}{3.75mm}
	\caption{Comparison results in terms of top-1 accuracy on ShapeNet dataset \cite{DBLP:journals/corr/ChangFGHHLSSSSX15}, where each incremental state has an increment of six classes, except for the last one with five new classes when $S=9$. }
	\scalebox{1.00}{
		\begin{tabular}{|c|ccccccccc|c|c|}
			\hline
			Incremental State $S$ & 1 & 2 & 3 & 4 & 5 & 6 & 7 & 8 & 9 & \multirow{2}{*}{Avg.} & \multirow{2}{*}{$\triangle$ (\%)} \\
			Number of Classes & 6 & 12 & 18 & 24 & 30 & 36 & 42 & 48 & 53 &  &  \\
			\hline
			LwF \cite{10.1007/978-3-319-46493-0_37}	& 96.3 & 86.8 & 78.5 & 68.3 & 60.7 & 52.4 & 45.1 & 42.6 & 39.5 & 63.4 & $\downarrow$20.3 \\
			iCaRL \cite{Rebuffi_2017_CVPR} & 96.7 & 88.4 & 82.1 & 74.9 & 68.5 & 62.3 & 56.9 & 51.3 & 44.6 & 69.5 & $\downarrow$14.2 \\
			DeeSIL \cite{deesil-eccv2018} & 97.1 & 90.2 & 84.3 & 76.5 & 73.7 & 65.6 & 57.3 & 53.6 & 47.2 & 71.7 & $\downarrow$12.0 \\
			EEIL \cite{Castro_2018_ECCV} & 97.3 & 91.8 & 86.4 & 79.5 & 73.1 & 67.3 & 63.4 & 57.1 & 51.6 & 74.2 & $\downarrow$9.5 \\
			IL2M \cite{Belouadah_2019_ICCV} & 97.5 & 91.4 & 86.7 & 79.8 & 75.6 & 71.8 & 69.1 & 64.8 & 61.4 & 77.6 & $\downarrow$6.1 \\
			DGMw \cite{Oleksiy_2019_CVPR} & 97.2 & 90.8 & 85.9 & 78.3 & 74.4 & 69.5 & 62.4 & 56.3 & 49.2 & 73.8 & $\downarrow$9.9 \\
			DGMa \cite{Oleksiy_2019_CVPR} & 97.2 & 91.6 & 85.1 & 77.9 & 73.2 & 68.5 & 62.8 & 55.4 & 48.7 & 73.4 & $\downarrow$10.3 \\
			BiC \cite{Wu_2019_CVPR} & 97.4 & 92.1 & 86.7 & 81.5 & 76.4 & 73.7 & 69.8 & 67.6 & 64.2 & 78.8 & $\downarrow$4.9  \\
			RPS-Net \cite{NIPS2019_9429} & 97.6 & 92.5 & 87.4 & 80.1 & 77.4 & 72.3 & 68.4 & 66.5 & 63.5 & 78.4 & $\downarrow$5.3 \\
			LUCIR \cite{Hou_2019_CVPR} + DDE \cite{Hu_20121_CVPR} & 97.6 & 93.1 & 88.7 & 82.1 & 78.5 & 74.5 & 71.2 & 67.9 & 64.8 & 79.8 & $\downarrow$3.9  \\
			LUCIR \cite{Hou_2019_CVPR} + GeoDL \cite{Christian2021MGeoCont} & 97.6 & 93.7 & 89.3 & 82.9 & 79.3 & 75.2 & 71.9 & 69.4 & 66.3 & 80.5 & $\downarrow$3.2 \\
			I3DOL \cite{I3DOL_Dong_AAAI2021} & 97.5 & 94.4 & 90.2 & 84.3 & 80.5 & 76.1 & 73.5 & 70.8 & 67.3 & 81.6 & $\downarrow$2.1 \\
			\hline
			\hline
			
			Ours-w/oCGR & 97.3 & 92.4 & 87.9 & 81.6 & 78.6 & 74.6 & 71.6 & 69.3 & 64.7 & 79.8 & $\downarrow$3.9  \\
			
			Ours-w/oCGA & 97.4 & 93.1 & 89.4 & 82.2 & 79.1 & 75.7 & 72.8 & 70.6 & 66.1 & 80.7 & $\downarrow$3.0 \\
			
			Ours-w/oWFC & \textcolor[rgb]{0.698,0.133,0.133}{\textbf{97.7}} & 93.7 & 90.5 & 82.6 & 80.8 & 76.1 & 73.4 & 71.5 & 66.8 & 81.5 & $\downarrow$2.2 \\
			
			Ours-w/oSFC & 97.6 & 94.2 & 91.3 & 84.8 & 81.4 & 77.6 & 74.3 & 72.8 & 67.9 & 82.4 & $\downarrow$1.3  \\
			
			Ours & 97.5 & \textcolor[rgb]{0.698,0.133,0.133}{\textbf{95.6}} & \textcolor[rgb]{0.698,0.133,0.133}{\textbf{92.4}} & \textcolor[rgb]{0.698,0.133,0.133}{\textbf{86.7}} & \textcolor[rgb]{0.698,0.133,0.133}{\textbf{83.1}} & \textcolor[rgb]{0.698,0.133,0.133}{\textbf{79.2}} & \textcolor[rgb]{0.698,0.133,0.133}{\textbf{76.0}} & \textcolor[rgb]{0.698,0.133,0.133}{\textbf{73.5}} & \textcolor[rgb]{0.698,0.133,0.133}{\textbf{69.4}} & \textcolor[rgb]{0.698,0.133,0.133}{\textbf{83.7}} & --- \\
			\hline					
		\end{tabular}
	} 	
	\label{tab:exp_shapenet54_dataset}
\end{table*}

Intuitively, Eq.~\eqref{eq:weight_fairness_compensation} encourages the average norm of $W_{\mr{new}}^l$ from new 3D classes to approximate $W_{\mr{old}}^l$ from old classes. 
Such design guarantees the prediction fairness among the old and new categories by adaptively adjusting the probabilities of new classes with $\frac{\mr{mean}(N_{\mr{old}})}{\mr{mean}(N_{\mr{new}})}$. Moreover, it can cooperate with the aforementioned critic network $\Gamma_c(\cdot)$ together to effectively address the catastrophic forgetting on old classes.

$\bullet$ \textbf{Score Fairness Compensation:}
In addition to the biased classifier $C(\cdot)$ in the training phase, the biased predictions during the inference phase are also non-negligible. To address this concern, we leverage the statistical information of old classes during the training phase to modify the prediction scores of new categories in the inference phase \cite{I3DOL_Dong_AAAI2021, Belouadah_2019_ICCV}. When adequate 3D objects of old classes are available, the predictions for them are more reliable. Thus, we record their initial statistical information for rectification when the old classes are initially learned in each incremental state. Then the rectified probability $C^s(f_g; \theta_{C})^k$ of the $k$-th category is written as: 
\begin{equation}
	\begin{split}
		C^s(f_g; \theta_{C})^k =
		\left\{
		\begin{aligned}		
			&C(f_g; \theta_{C})^k \cdot \frac{\psi_{s_i}(k)}{\psi_{s}(k)} \cdot \frac{\psi(s)}{\psi(s_i)}, \text{if new category},  \\
			&C(f_g; \theta_{C})^k, \text{otherwise}, \\
		\end{aligned} 					
		\right.											
	\end{split}							
	\label{eq:score_fairness_compensation^a}	
\end{equation}
where $\psi_{s_i}(k)$ and $\psi_s(k)$ respectively represent the mean scores classified as the $k$-th categories in the initial $s_i$-th state and the current $s$-th state. Here, the initial $s_i$-th state indicates that all 3D objects of the $k$-th category are available. $\psi(s_i)$ and $\psi(s)$ denote the average scores of new categories in the $s_i$-th and $s$-th incremental states. We need to mention that only if the given 3D objects are initially predicted as the new classes, Eq.~\eqref{eq:score_fairness_compensation^a} applies rectification to their predicted scores. Intuitively, Eq.~\eqref{eq:score_fairness_compensation^a} can guarantee the test fairness among the old and new categories of 3D objects by adaptively adjusting the probabilities of new classes with $\frac{\psi_{s_i}(k)}{\psi_{s}(k)} \cdot \frac{\psi(s)}{\psi(s_i)}$ in the inference phase.

\subsection{Implementation Details}
For the network configuration, as shown in Table~\ref{tab:network_implementation}, we provide the detailed architecture description of the proposed InOR-Net model. Specifically, we use PointNet \cite{Qi_2017_CVPR} as the encoder $E(\cdot)$ to extract the low-level representations of point clouds, and employ a four-layer fully-connected network as the classifier $C(\cdot)$. The channels of $C(\cdot)$ are set as $\{1024, 512, 256, \mr{number~of~classes}\}$ in this paper.  
The Adam optimizer with the initial learning rate as $0.001$ is utilized to optimize our model, and its weight decay is set as $0.0005$. 
The proposed InOR-Net model is implemented via PyTorch, and we follow the parameter initialization manner\footnote{https://github.com/fxia22/pointnet.pytorch} proposed in \cite{Qi_2017_CVPR} to initialize our model.  
In Section~\ref{sec:category_guided_geometric_reasoning}, we empirically set the number of local geometric structures as $L=64$ via conducting the parameter experiments in Section~\ref{sec: parameters_analysis}. The mini-batch size $T_B$ is set as 64 for all benchmark datasets. 
Inspired by \cite{Christian2021MGeoCont}, we utilize the similar method (\emph{i.e.}, herding strategy) to select the exemplar set $M$. Overall, the formulation $\mathcal{L}_{\mr{obj}}$ of our InOR-Net model is written as follows: 
\begin{align}
\mathcal{L}_{\mr{obj}} = \mathcal{L}_{\mr{clc}} + \mathcal{L}_{\mr{reg}} + \lambda_1\mathcal{L}_{\mr{cri}} + \lambda_2\mathcal{L}_{\mr{cst}},
\label{eq:overall_training_objective}
\end{align}
where $\mathcal{L}_{\mr{clc}} \!=\! \mathbb{E}_{(x_i^s, y_i^s)\in D_s\cup M}[-\! \sum\nolimits_{k=1}^{K_p+K_s}\!(y_i^s)^k\!\log(C(f_g; \theta_{C})^k)]$ is the cross-entropy loss for classification. $\lambda_1, \lambda_2\geq0$ are the balanced weights, which are empirically set as $\lambda_1=0.01, \lambda_2=0.1$. We also summarize the optimization procedure of our InOR-Net model, as shown in \textbf{Algorithm \ref{alg:InORNet}}.

\textbf{Inference:} For performance evaluation, we directly forward the 3D object into the encoder $E(\cdot)$ to extract $f_m$ via category-guided geometric reasoning, and obtain $f_p$ via geometric attention network $\Gamma_a(\cdot)$. After performing max-pooling on $f_p$, we get the global feature $f_g$, and forward it into the classifier $C(\cdot)$ to compute softmax probabilities that are rectified via Eq.~\eqref{eq:score_fairness_compensation^a} for the final 3D object recognition.

\begin{table*}[t]
	\centering
	\setlength{\tabcolsep}{3.29mm}
	\caption{Comparison results in terms of top-1 accuracy on ModelNet dataset \cite{7298801}, where each incremental state has an increment of four new classes when $S=10$. }
	\scalebox{1.0}{
		\begin{tabular}{|c|cccccccccc|c|c|}
			\hline
			Incremental State $S$ & 1 & 2 & 3 & 4 & 5 & 6 & 7 & 8 & 9 & 10 & \multirow{2}{*}{Avg.} & \multirow{2}{*}{$\triangle$ (\%)} \\
			Number of Classes & 4 & 8 & 12 & 16 & 20 & 24 & 28 & 32 & 36 & 40 & & \\
			\hline
			LwF \cite{10.1007/978-3-319-46493-0_37}	& 96.5 & 87.2 & 77.5 & 70.6 & 62.3 & 56.8 & 44.7 & 39.4 & 36.1 & 31.5 & 60.3 & $\downarrow$26.7 \\
			iCaRL \cite{Rebuffi_2017_CVPR} & 96.8 & 90.4 & 83.6 & 78.3 & 72.5 & 67.3 & 59.6 & 53.1 & 47.8 & 39.6 & 68.9 & $\downarrow$18.1 \\
			DeeSIL \cite{deesil-eccv2018} & 97.7 & 91.5 & 85.4 & 80.5 & 74.4 & 71.8 & 65.3 & 58.7 & 52.4 & 43.7 & 72.1 & $\downarrow$14.9 \\
			EEIL \cite{Castro_2018_ECCV} & 97.6 & 93.8 & 87.5 & 81.6 & 78.2 &  74.7 & 69.2 & 62.4 & 56.8 & 48.1 & 75.0 & $\downarrow$12.0 \\
			IL2M \cite{Belouadah_2019_ICCV} & 97.8 & 95.1 & 89.4 & 85.7 & 83.8 & 82.2 & 78.4 & 72.8 & 67.9 & 57.6 & 81.1 & $\downarrow$5.9 \\
			DGMw \cite{Oleksiy_2019_CVPR} & 97.5 & 93.2 & 86.4 & 82.5 & 80.1 & 78.4 & 73.6 & 65.3 & 61.5 & 53.4 & 77.2 & $\downarrow$9.8 \\
			DGMa \cite{Oleksiy_2019_CVPR} & 97.5 & 93.4 & 84.7 & 81.8 & 79.5 & 77.8 & 74.1 & 67.4 & 60.8 & 51.5 & 76.8 & $\downarrow$10.2 \\
			BiC \cite{Wu_2019_CVPR} & 97.8 & 95.5 & 88.5 & 86.9 & 84.3 & 83.1 & 79.3 & 74.2 & 70.7 & 59.2 & 82.0 & $\downarrow$5.0 \\
			RPS-Net \cite{NIPS2019_9429} & 97.7 & 94.6 & 90.3 & 88.2 & 86.7 & 82.5 & 78.0 & 73.6 & 68.4 & 58.3 & 81.7 & $\downarrow$5.3 \\
			
			LUCIR \cite{Hou_2019_CVPR} + DDE \cite{Hu_20121_CVPR} & 97.9 & 96.1 & 91.4 & 88.9 & 87.5 & 84.6 & 80.4 & 75.1 & 71.4 & 59.8 & 83.3 & $\downarrow$3.7 \\
			LUCIR \cite{Hou_2019_CVPR} + GeoDL \cite{Christian2021MGeoCont} & 97.8 & 96.5 & 92.2 & 90.3 & 88.4 & 85.8 & 81.3 & 76.3 & 72.2 & 60.3 & 84.1 & $\downarrow$2.9 \\
			I3DOL \cite{I3DOL_Dong_AAAI2021} & 98.1 & 97.0 & 93.4 & 91.1 & 89.7 & 88.2 & 83.5 & 77.8 & 73.1 & 61.5 & 85.3 & $\downarrow$1.7 \\
			\hline
			\hline
			Ours-w/oCGR & 97.8 & 94.6 & 90.9 & 89.7 & 87.4 & 85.8 & 81.2 & 74.6 & 71.3 & 58.4 & 83.2 & $\downarrow$3.8  \\
			
			Ours-w/oCGA & 98.1 & 95.6 & 92.2 & 91.8 & 88.3 & 86.7 & 82.9 & 75.3 & 72.1 & 60.7 & 84.4 & $\downarrow$2.6  \\
			
			Ours-w/oWFC & 98.0 & 96.4 & 92.8 & 92.5 & 89.4 & 87.3 & 84.6 & 77.2 & 72.6 & 61.1 & 85.3 & $\downarrow$1.8 \\
			
			Ours-w/oSFC & \textcolor[rgb]{0.698,0.133,0.133}{\textbf{98.2}} & 96.7 & 93.1 & 93.2 & 90.6 & 88.1 & 83.2 & 78.4 & 73.5 & 62.8 & 85.8 & $\downarrow$1.2 \\
			
			Ours & 98.1 & \textcolor[rgb]{0.698,0.133,0.133}{\textbf{97.5}} & \textcolor[rgb]{0.698,0.133,0.133}{\textbf{95.6}} & \textcolor[rgb]{0.698,0.133,0.133}{\textbf{93.7}} & \textcolor[rgb]{0.698,0.133,0.133}{\textbf{91.4}} & \textcolor[rgb]{0.698,0.133,0.133}{\textbf{90.3}} & \textcolor[rgb]{0.698,0.133,0.133}{\textbf{85.9}} & \textcolor[rgb]{0.698,0.133,0.133}{\textbf{79.2}} & \textcolor[rgb]{0.698,0.133,0.133}{\textbf{74.6}} & \textcolor[rgb]{0.698,0.133,0.133}{\textbf{63.9}} &  \textcolor[rgb]{0.698,0.133,0.133}{\textbf{87.0}} & --- \\
			\hline					
		\end{tabular}
	} 	
	\label{tab:exp_modelnet40_dataset}
\end{table*}

\begin{table*}[t]
	\centering
	\setlength{\tabcolsep}{3.75mm}
	\caption{Comparison results in terms of top-1 accuracy on ScanNet dataset \cite{8099744}, where each incremental state has an increment of two classes, except for the last one with only one new class when $S=9$. }
	\scalebox{1.00}{
		\begin{tabular}{|c|ccccccccc|c|c|}
			\hline
			Incremental State $S$ & 1 & 2 & 3 & 4 & 5 & 6 & 7 & 8 & 9 & \multirow{2}{*}{Avg.} & \multirow{2}{*}{$\triangle$ (\%)} \\
			Number of Classes & 2 & 4 & 6 & 8 & 10 & 12 & 14 & 16 & 17 &  & \\
			\hline
			LwF \cite{10.1007/978-3-319-46493-0_37}	& 92.2 & 74.8 & 60.3 & 48.2 & 41.6 & 37.3 & 35.7 & 33.5 & 31.8 & 53.1 & $\downarrow$19.1 \\
			iCaRL \cite{Rebuffi_2017_CVPR} & 92.4 & 78.7 & 67.4 & 59.7 & 52.5 & 48.2 & 43.5 & 39.9 & 36.3 & 56.0 & $\downarrow$16.2 \\
			DeeSIL \cite{deesil-eccv2018} & 92.6 & 80.1 & 71.5 & 63.3 & 57.3 & 52.8 & 48.6 & 45.2 & 43.7 & 63.1 & $\downarrow$9.1 \\
			EEIL \cite{Castro_2018_ECCV} & 92.7 & 83.4 & 75.6 & 72.6 & 58.7 & 55.4 & 52.3 & 49.4 & 45.7 & 65.1 & $\downarrow$7.1 \\
			IL2M \cite{Belouadah_2019_ICCV} & 92.9 & 84.4 & 77.3 & 70.1 & 60.8 & 56.7 & 54.1 & 52.6 & 48.3 & 66.7 & $\downarrow$5.5 \\
			DGMw \cite{Oleksiy_2019_CVPR} & 92.5 & 82.6 & 67.1 & 61.8 & 56.3 & 53.2 & 50.8 & 47.5 & 43.8 & 63.6 & $\downarrow$8.6 \\
			DGMa \cite{Oleksiy_2019_CVPR} & 92.5 & 82.2 & 67.8 & 60.2 & 56.6 & 52.7 & 50.4 & 48.1 & 44.7 & 63.7 & $\downarrow$8.5 \\
			BiC \cite{Wu_2019_CVPR} & 92.8 & 84.2 & 77.5 & 70.3 & 60.6 & 57.2 & 54.3 & 52.4 & 48.5 & 66.8 & $\downarrow$5.4  \\
			RPS-Net \cite{NIPS2019_9429} & 92.9 & 84.8 & 77.1 & 70.7 & 61.2 & 57.6 & 55.4 & 53.3 & 49.1 & 67.3 & $\downarrow$4.9 \\
			
			LUCIR \cite{Hou_2019_CVPR} + DDE \cite{Hu_20121_CVPR} & 92.9 & 85.6 & 78.2 & 72.1 & 61.8 & 58.4 & 56.6 & 54.0 & 49.8 & 67.7 & $\downarrow$4.5  \\
			LUCIR \cite{Hou_2019_CVPR} + GeoDL \cite{Christian2021MGeoCont} & 92.9 & 86.3 & 78.9 & 74.0 & 62.5 & 59.2 & 57.1 & 55.3 & 50.7 & 68.5 & $\downarrow$3.7 \\
			I3DOL \cite{I3DOL_Dong_AAAI2021} & 93.2 & 87.2 & 80.5 & 77.8 & 64.3 & 61.9 & 58.2 & 56.8 & 52.1 & 70.2 & $\downarrow$2.0 \\
			\hline
			\hline

			Ours-w/oCGR & 92.9 & 84.8 & 77.3 & 75.2 & 64.4 & 62.0 & 57.1 & 54.1 & 50.7 & 68.7 & $\downarrow$3.5 \\
			
			Ours-w/oCGA & 93.2 & 85.3 & 78.1 & 76.9 & 65.1 & 61.8 & 57.4 & 54.5 & 51.8 & 69.3 & $\downarrow$2.9  \\
			
			Ours-w/oWFC & \textcolor[rgb]{0.698,0.133,0.133}{\textbf{93.3}} & 86.4 & 79.6 & 77.1 & 65.4 & 62.5 & 58.5 & 55.1 & 52.6 & 70.1 & $\downarrow$2.1 \\
			
			Ours-w/oSFC & 93.2 & 87.0 & 80.8 & 77.5 & 66.1 & 62.8 & 58.2 & 56.7 & 53.1 & 70.6 & $\downarrow$1.6 \\ 
			
			Ours & 93.2 & \textcolor[rgb]{0.698,0.133,0.133}{\textbf{88.7}} & \textcolor[rgb]{0.698,0.133,0.133}{\textbf{82.6}} & \textcolor[rgb]{0.698,0.133,0.133}{\textbf{79.4}} & \textcolor[rgb]{0.698,0.133,0.133}{\textbf{67.9}} & \textcolor[rgb]{0.698,0.133,0.133}{\textbf{64.0}} & \textcolor[rgb]{0.698,0.133,0.133}{\textbf{60.6}} & \textcolor[rgb]{0.698,0.133,0.133}{\textbf{58.3}} & \textcolor[rgb]{0.698,0.133,0.133}{\textbf{54.8}} & \textcolor[rgb]{0.698,0.133,0.133}{\textbf{72.2}} & --- \\
			\hline					
		\end{tabular}
	} 	
	\label{tab:exp_scannet17_dataset}
\end{table*}

\begin{figure*}[t]
	\centering
	\includegraphics[width =516pt, height =138pt] {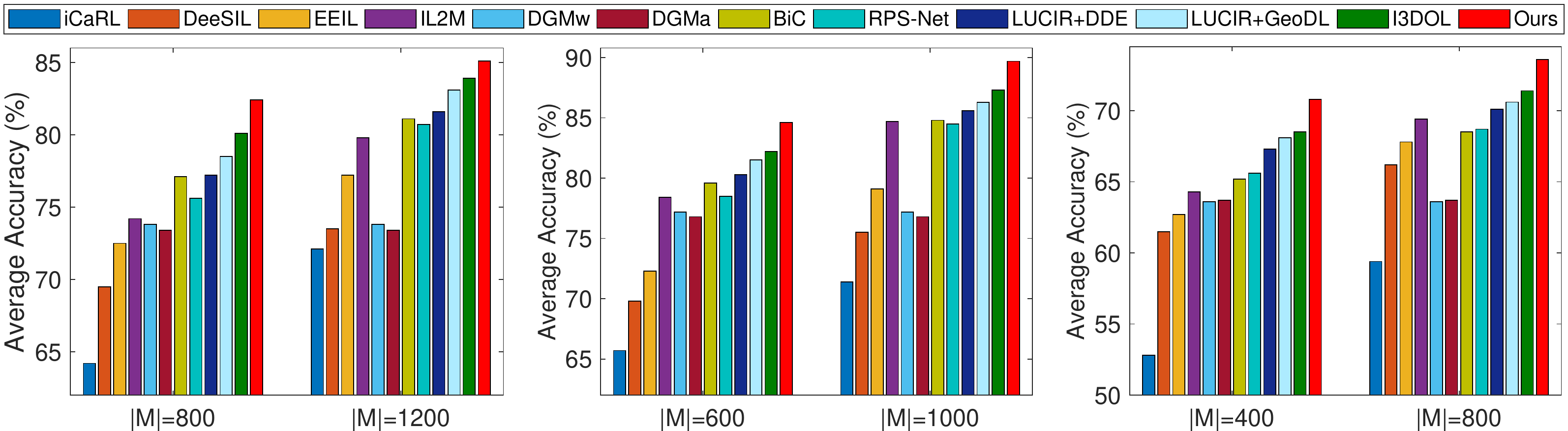}
	\vspace{-20pt}
	\caption{Qualitative analysis of the exemplar set $M$ on ShapeNet \cite{DBLP:journals/corr/ChangFGHHLSSSSX15} (left), ModelNet \cite{7298801} (middle) and ScanNet \cite{8099744} (right) datasets. } 
	\label{fig:effect_exemplar_set}
	\vspace{-10pt}
\end{figure*}

\begin{figure*}[t]
	\centering
	\includegraphics[width =516pt, height =165pt] {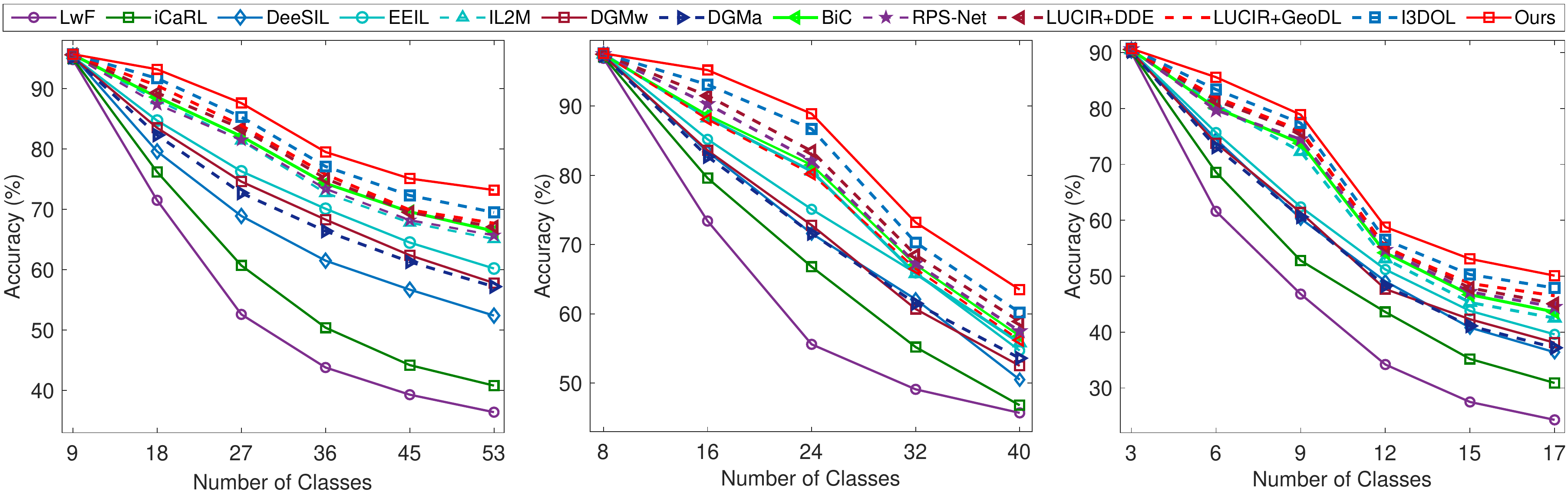}
	\vspace{-20pt}
	\caption{Qualitative analysis of the total incremental states $S$ on ShapeNet \cite{DBLP:journals/corr/ChangFGHHLSSSSX15} (left, $S=6, |M|=1000$), ModelNet \cite{7298801} (middle, $S=5, |M|=800$) and ScanNet \cite{8099744} (right, $S=6, |M|=600$) datasets. }
	\label{fig:effect_incremental_states}
	\vspace{-10pt}
\end{figure*}

\section{Experiments}

\subsection{Datasets and Evaluation Metric}

$\bullet$ \textbf{ShapeNet} \cite{DBLP:journals/corr/ChangFGHHLSSSSX15} is composed of 35037 CAD samples for training and 5053 CAD samples for validation, which are collected from online repositories. We use 53 categories in the comparison experiments, and select 1000 CAD samples to be stored in the exemplar set $M$. The total incremental states $S$ is set as 9, and each incremental state has an increment of six classes, except for the last one with five new classes. 

$\bullet$ \textbf{ModelNet} \cite{7298801} with 40 different CAD categories contains 9843 samples for training and 2468 samples for evaluation. All CAD samples are clean models. The exemplar set $M$ stores 800 CAD samples. We define the total incremental states $S$ as 10, and each incremental state will collect four new classes. 

$\bullet$ \textbf{ScanNet} \cite{8099744} consists of 17 different classes collected from the scanned and reconstructed scenes. When compared with ShapeNet \cite{DBLP:journals/corr/ChangFGHHLSSSSX15} and ModelNet \cite{DBLP:journals/corr/ChangFGHHLSSSSX15}, it is a more challenging dataset due to many noisy geometric structures. The training and test sets respectively have 12060 and 3416 samples. We store 600 CAD samples in the exemplar set $M$, and define total incremental states $S=9$, where each incremental state has an increment of two classes, except for the last one with only one incremental class. Moreover, some example samples of three benchmark datasets are visualized in Fig.~\ref{fig:example_samples}.

As for the selection of exemplar set $M$ and incremental states $S$, we follow the standard settings of class-incremental learning proposed in \cite{Rebuffi_2017_CVPR, Castro_2018_ECCV, Wu_2019_CVPR, NIPS2019_9429, Christian2021MGeoCont, Hu_20121_CVPR}, and utilize the same experimental settings to compare with other state-of-the-art baseline methods for fair comparisons. Specifically, considering the number of 3D object classes in each incremental state, we select the number of total incremental states to be around $10$ for more new categories in each incremental task, while ensuring that the number of incremental states $S$ is as large as possible. Besides, the number of exemplars from $M$ satisfies $|M|/K_p\ll n_s/K_s$, and we empirically set the number of $M$ as $1000, 800, 600$ for ShapeNet \cite{DBLP:journals/corr/ChangFGHHLSSSSX15}, ModelNet \cite{7298801} and ScanNet \cite{8099744} respectively.

\textbf{Evaluation Metric:}
Following other baseline methods \cite{Rebuffi_2017_CVPR, Wu_2019_CVPR, Hu_20121_CVPR, I3DOL_Dong_AAAI2021}, we utilize the top-1 accuracy \cite{Ceri2013, wei2021incremental} as the evaluation metric to conduct comparison experiments.

\subsection{Comparison Experiments}
The comparison experiments on ShapeNet \cite{DBLP:journals/corr/ChangFGHHLSSSSX15}, ModelNet \cite{7298801}, ScanNet \cite{8099744} are shown in Tables~\ref{tab:exp_shapenet54_dataset}, \ref{tab:exp_modelnet40_dataset} and \ref{tab:exp_scannet17_dataset}. For a fair comparison, all baseline methods utilize PointNet \cite{Qi_2017_CVPR} to obtain local feature of point cloud, and are trained with the same data augmentation mechanism in \cite{Qi_2017_CVPR}. According to the results, we observe our InOR-Net model outperforms the conference version I3DOL \cite{I3DOL_Dong_AAAI2021} about $1.7\%\sim2.1\%$ (average accuracy) on three benchmark datasets. The substantial extensions about exploring 3D geometric characteristics and alleviating catastrophic forgetting promote the object recognition ability of our InOR-Net model against the conference method \cite{I3DOL_Dong_AAAI2021}. Moreover, the average accuracy of Ours significantly outperforms other comparison methods \cite{Rebuffi_2017_CVPR, Castro_2018_ECCV, Belouadah_2019_ICCV, Oleksiy_2019_CVPR, Wu_2019_CVPR, 10.1007/978-3-319-46493-0_37, NIPS2019_9429, Christian2021MGeoCont, Hu_20121_CVPR} by a large margin about $2.9\%\sim26.7\%$. It illustrates that our model has more advantages than 2D methods in addressing 3D class-incremental learning. Specifically, when compared with them, our model is effective to explore distinctive 3D characteristics within each class under the category information as guidance via the category-guided geometric reasoning. Our InOR-Net model also considers different contributions of 3D characteristics to address the forgetting via critic-induced geometric attention. The recognition results of our InOR-Net model are better than knowledge distillation based approaches \cite{Castro_2018_ECCV, Rebuffi_2017_CVPR, Wu_2019_CVPR, NIPS2019_9429, Hu_20121_CVPR, Christian2021MGeoCont}, since the dual adaptive fairness compensations could effectively alleviate catastrophic forgetting by correcting prediction bias among old and new categories in the training and test stages.

\subsection{Ablation Studies}
To demonstrate the necessity of proposed modules in our InOR-Net model, we introduce detailed ablation studies on three benchmark datasets, as shown in Tables~\ref{tab:exp_shapenet54_dataset}, \ref{tab:exp_modelnet40_dataset} and \ref{tab:exp_scannet17_dataset}. Ours-w/oCGR, Ours-w/oCGA, Ours-w/oWFC and Ours-w/oSFC represent the performance of our model without using the category-guided geometric reasoning (CGR), critic-induced geometric attention (CGA), weight fairness compensation (WFC) and score fairness compensation (SFC).

Tables~\ref{tab:exp_shapenet54_dataset}, \ref{tab:exp_modelnet40_dataset} and \ref{tab:exp_scannet17_dataset} show that the average accuracy degrades $1.2\%\sim3.9\%$ when any component of our model is absent. It verifies that the proposed modules cooperate well to address 3D class-incremental learning. Specifically, the lack of dual adaptive fairness compensations causes the degradation about $1.2\%\sim2.2\%$ average accuracy, which validates the effectiveness to alleviate the forgetting via correcting the prediction bias. When removing the category-guided geometric reasoning, the average accuracy of our model decreases $3.5\%\sim3.9\%$. The worse performance explains its importance to explore unique 3D characteristics within each class. Ours-w/oCGA performs worse than Ours about $2.6\%\sim3.0\%$. It validates the effectiveness of critic-induced geometric attention to highlight distinctive 3D characteristics.

\begin{figure*}[t]
	\begin{minipage}[t]{0.331\linewidth}
		\centering
		\includegraphics[height=140pt, width=168pt] {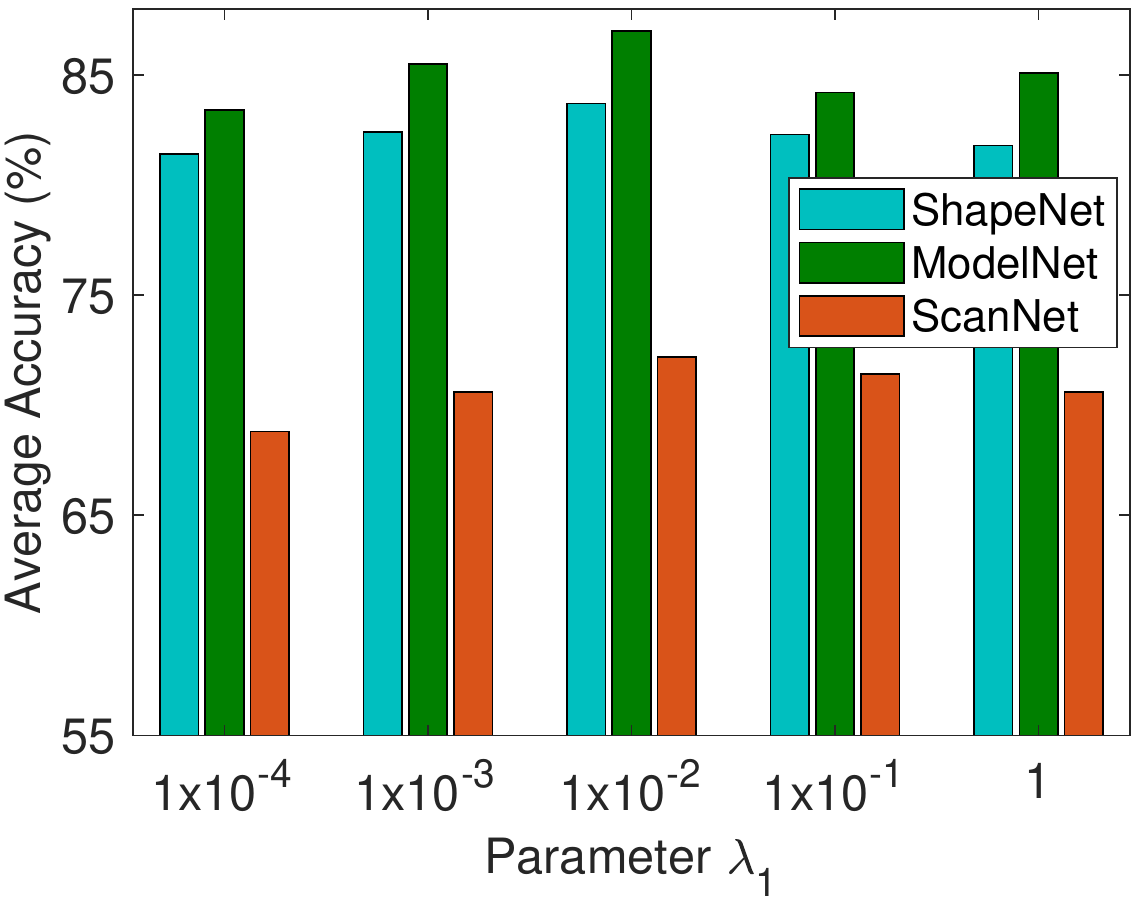}
		\\ (a) When $\lambda_2=0.1, L=64$
	\end{minipage}
	\begin{minipage}[t]{0.331\linewidth}
		\centering
		\includegraphics[height=140pt, width=168pt] {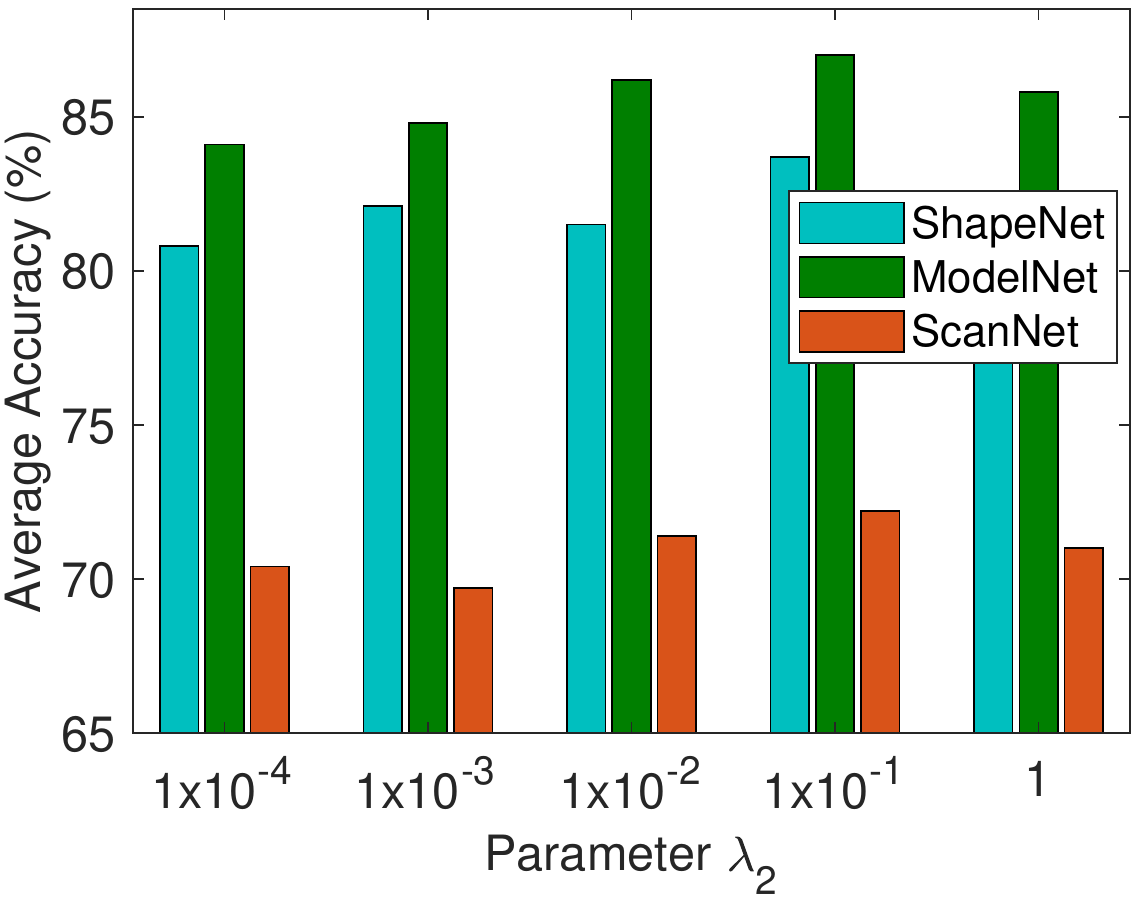}
		\\~~ (b) When $\lambda_1=0.01, L=64$
	\end{minipage}
	\begin{minipage}[t]{0.331\linewidth}
		\centering
		\includegraphics[height=140pt, width=168pt] {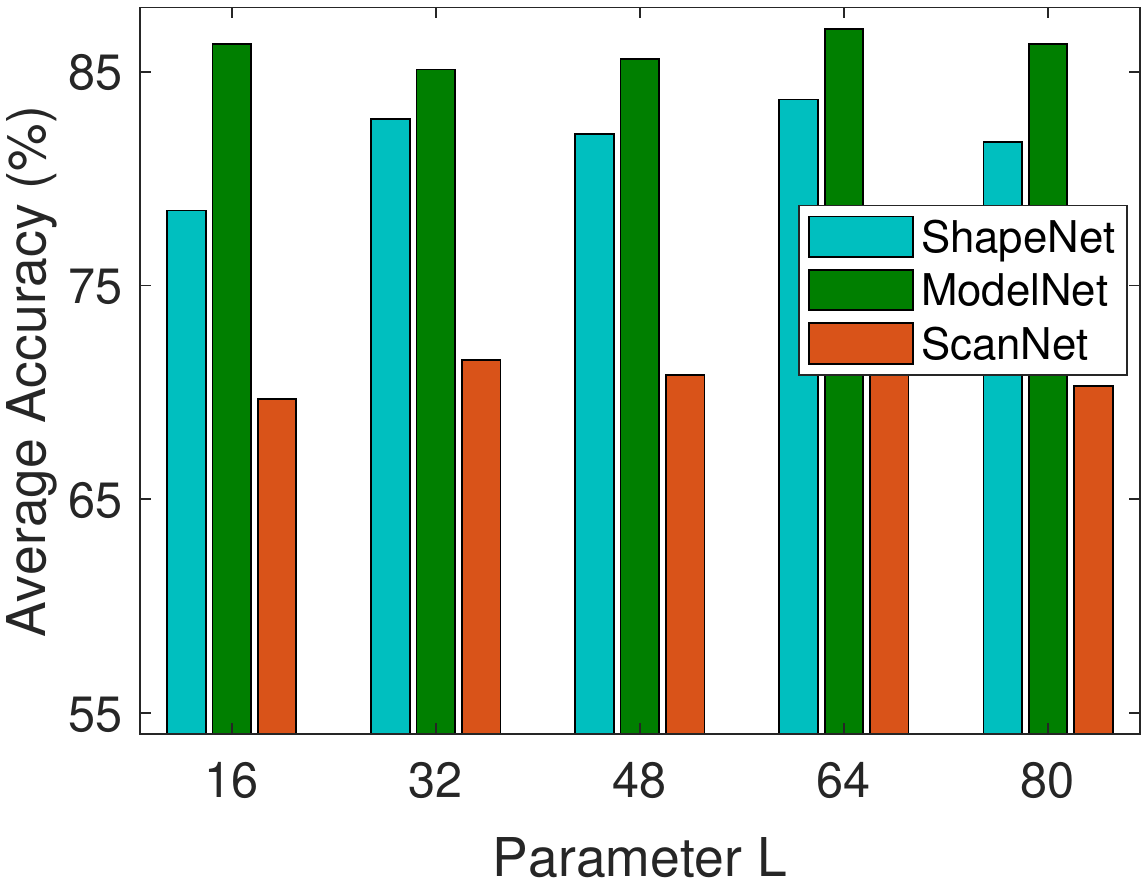}
		\\~~ (c) When $\lambda_1=0.01, \lambda_2=0.1$
	\end{minipage}
	\vspace{-10pt}
	\caption{Investigation experiments about parameters $\lambda_1$ (left), $\lambda_2$ (middle) and $L$ (right) on three benchmark datasets. }
	\label{fig:parameters_analysis}
	\vspace{-5pt}
\end{figure*}

\begin{figure*}[t]
	\begin{minipage}[t]{0.331\linewidth}
		\centering
		\includegraphics[trim = 0mm 0mm 0mm 0mm, clip, height=145pt, width=160pt]{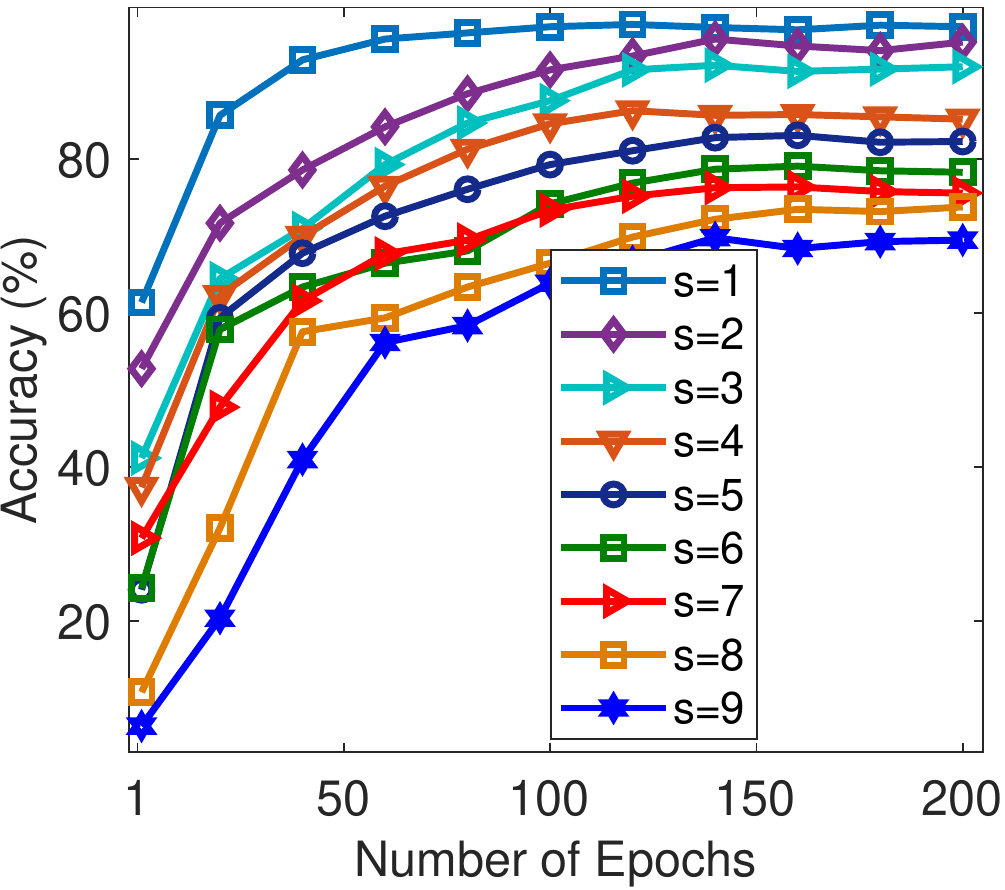}
	\end{minipage}
	\begin{minipage}[t]{0.331\linewidth}
		\centering
		\includegraphics[trim = 0mm 0mm 0mm 0mm, clip, height=145pt, width=160pt]{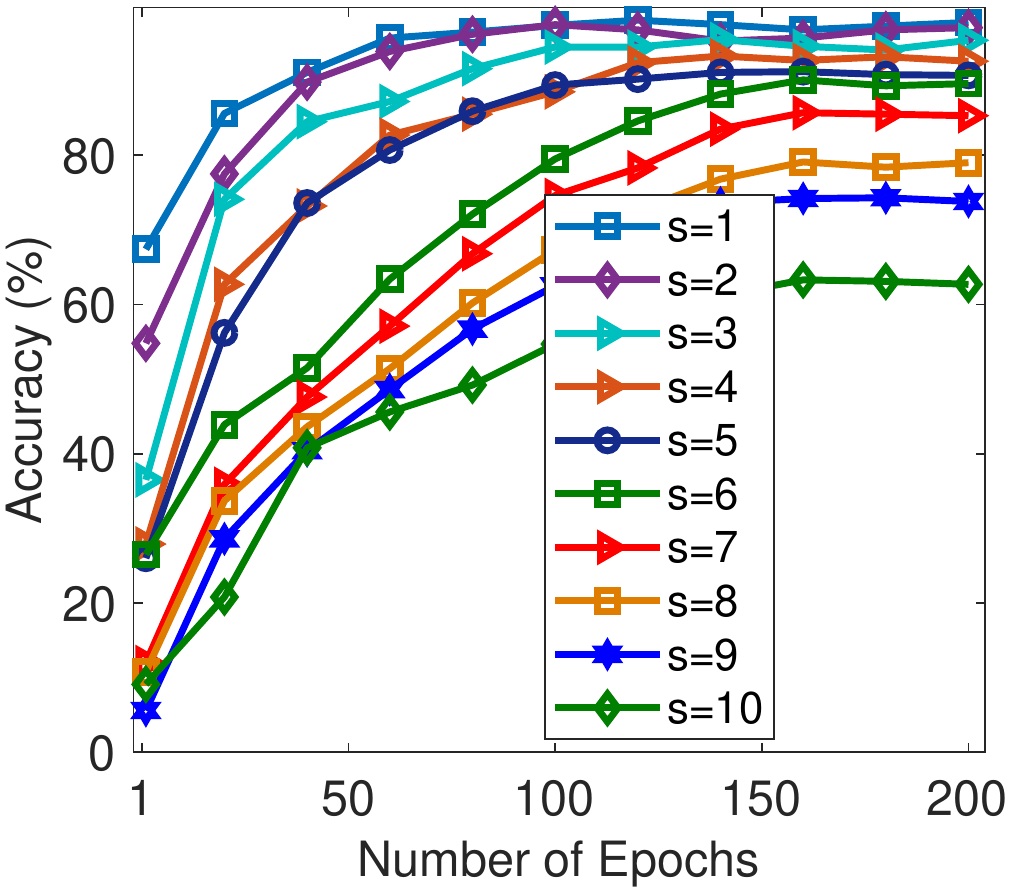}
	\end{minipage}
	\begin{minipage}[t]{0.331\linewidth}
		\centering
		\includegraphics[trim = 0mm 0mm 0mm 0mm, clip, height=145pt, width=160pt]{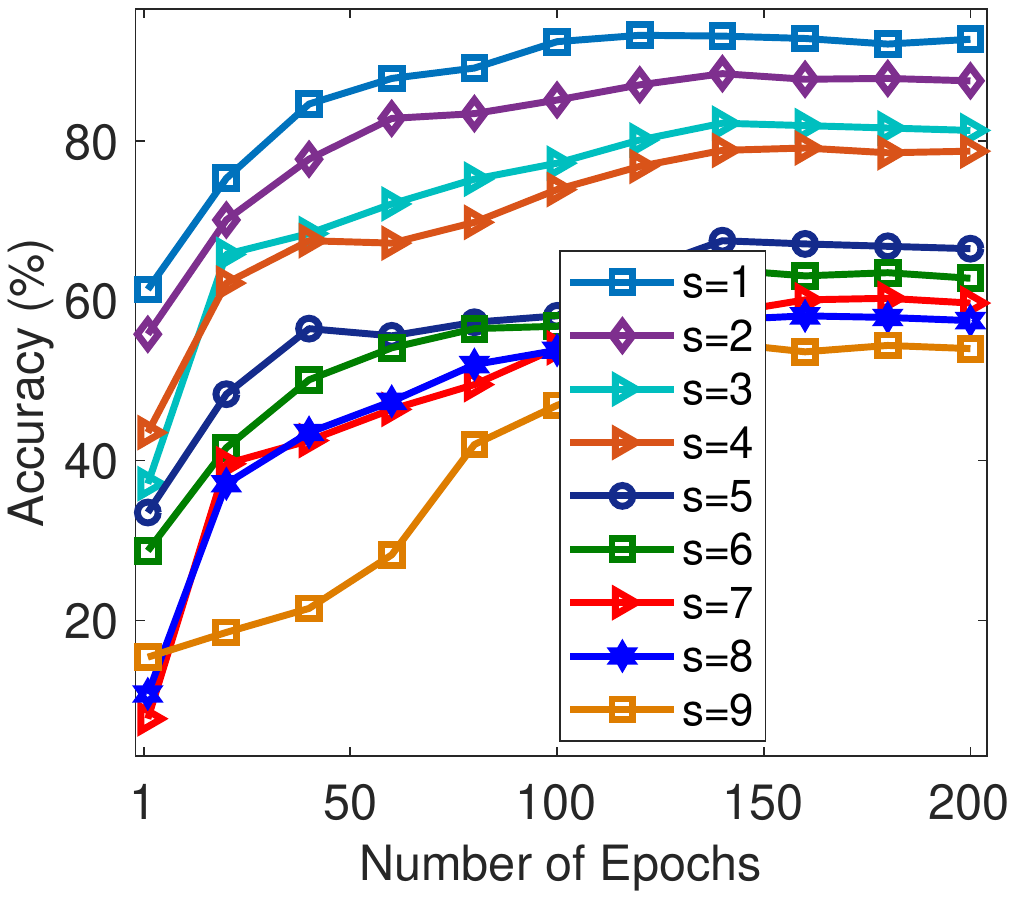}
	\end{minipage}
	\vspace{-20pt}
	\caption{Qualitative analysis of model convergence on ShapeNet \cite{DBLP:journals/corr/ChangFGHHLSSSSX15} (left), ModelNet \cite{7298801} (middle) and ScanNet \cite{8099744} (right) datasets, where $s=1, 2, \cdots, S$ denotes the convergence curves of different incremental states. } 
	\label{fig:convergence_analysis}
	\vspace{-5pt}
\end{figure*}

\subsection{Qualitative Analysis of Exemplar Set}
This section investigates the effect of exemplar set $M$ on model performance. As depicted in Fig.~\ref{fig:effect_exemplar_set}, we present comparison experiments on benchmark datasets when setting different sizes of the exemplar set $M$. Specifically, we set the sample number of $M$ as $\{800, 1200\}$, $\{600, 1000\}$ and $\{400, 800\}$ for ShapeNet \cite{DBLP:journals/corr/ChangFGHHLSSSSX15}, ModelNet \cite{7298801} and ScanNet \cite{8099744} datasets respectively. According to Fig.~\ref{fig:effect_exemplar_set}, we can observe that our model still outperforms all comparison baselines when setting a small sample number of exemplar set $M$, which further validates the superiority of our model. This observation also illustrates the effectiveness of our model against other comparison methods, when addressing the forgetting on old classes of 3D objects. Moreover, training our model with more exemplars significantly facilitates to alleviate the catastrophic forgetting by exploring distinctive 3D geometric characteristics and addressing the class imbalance.

\subsection{Qualitative Analysis of Incremental States}
As depicted in Fig.~\ref{fig:effect_incremental_states}, we introduce the comparison experiments between our model and other baseline methods on benchmark datasets, when setting different incremental states $S$ with Tables~\ref{tab:exp_shapenet54_dataset}, \ref{tab:exp_modelnet40_dataset} and \ref{tab:exp_scannet17_dataset}. From the depicted curves in Fig.~\ref{fig:effect_incremental_states}, we observe that all comparison methods perform worse than Ours, even though each incremental state has more new 3D classes. It validates the generalization of our model to alleviate the catastrophic forgetting under different experimental configurations.  
Compared with the conference version I3DOL \cite{I3DOL_Dong_AAAI2021}, our model could effectively quantify the contributions of distinctive 3D characteristics within each class via the critic-induced geometric attention, and address the forgetting caused by class imbalance via the dual adaptive fairness compensations. These substantial extensions lead to the performance improvements of our model. Besides, the recognition results of our model are significantly better than other 2D baseline approaches \cite{Castro_2018_ECCV, Belouadah_2019_ICCV, Oleksiy_2019_CVPR, NIPS2019_9429, Christian2021MGeoCont, Hu_20121_CVPR}, since the category-guided geometric reasoning captures distinctive 3D properties within each class via using category information as learning guidance.

\subsection{Qualitative Analysis of Parameters $\{\lambda_1, \lambda_2, L\}$} \label{sec: parameters_analysis}
This subsection presents the parameter experiments about $\lambda_1, \lambda_2$ in Eq.~\eqref{eq:overall_training_objective} and $L$ in Section \ref{sec:category_guided_geometric_reasoning} on three benchmark datasets, by tuning them in range of $\{1\times 10^{-4}, 1\times 10^{-3}, 1\times 10^{-2}, 1\times 10^{-1}, 1\}$ and $\{16, 32, 48, 64, 80\}$, respectively. As shown in Fig.~\ref{fig:parameters_analysis}, when we set $\lambda_1=0.01, \lambda_2=0.1, L=64$, our model achieves the optimal average accuracy, and we utilize these parameter values in comparison experiments. Fig.~\ref{fig:parameters_analysis} also shows the performance of our model has great stability even though $\{\lambda_1, \lambda_2, L\}$ have a wide selection range. 
Furthermore, when $L=64$ and fixing other parameters $\{\lambda_1, \lambda_2\}$, our model has the best performance on benchmark datasets. It validates each point cloud could be characterized well by 64 local geometric structures. Besides, the performance of our model degrades when setting an inappropriate value for $L$. It illustrates that more local geometric structures can bring more noise information, while less local geometric structures may lose some informative knowledge.  
Our model can capture unique 3D characteristics within each class via the category-guided geometric reasoning (\emph{i.e.}, $\lambda_2\mathcal{L}_{\mr{cst}}$). $\lambda_1\mathcal{L}_{\mr{cri}}$ provides the positive supervision to guide $\Gamma_a(\cdot)$ maximize the gain of unique 3D characteristics via tuning the parameter $\lambda_1$.

\subsection{Convergence Analysis}
We introduce the convergence investigation of our InOR-Net model in this subsection, as shown in Fig.~\ref{fig:convergence_analysis}. From the convergence curves, we can observe that when the number of training epoches is about 140, the accuracy performance of our InOR-Net model converges to a stable value on three benchmark datasets. Moreover, the stable performances across incremental states verify our model can recognize new categories of 3D objects consecutively while tackling the forgetting on old categories. It also demonstrates that the proposed modules cooperate well to address 3D class-incremental learning.

\subsection{Qualitative Analysis of Time and Memory Complexities}
As shown in Table~\ref{tab:time_memory_comparison}, we conduct time and computational parameters comparisons between our InOR-Net model and other approaches on ModelNet \cite{7298801} under the same settings. The training time (h) denotes the averaged model convergence time (hour) across all incremental states, the test time (s) shows the inference time cost (second) of mini-batch data, and the \#Parameters (M) represents computational network parameters of the model trained with TITAN XP GPU. As introduced in Table~\ref{tab:time_memory_comparison}, we conclude that our proposed InOR-Net model significantly outperforms \cite{Hou_2019_CVPR, Christian2021MGeoCont, Oleksiy_2019_CVPR, Hu_20121_CVPR} by a large margin, and is comparable with \cite{I3DOL_Dong_AAAI2021, NIPS2019_9429, Wu_2019_CVPR, Rebuffi_2017_CVPR, Castro_2018_ECCV, Belouadah_2019_ICCV} in terms of training time, test time and network parameters. Our model sacrifices marginal computational time and parameters, but achieves significant performance improvement (see Tables~\ref{tab:exp_shapenet54_dataset}, \ref{tab:exp_modelnet40_dataset} and \ref{tab:exp_scannet17_dataset}) than competing approaches, which is acceptable in real-world applications.

\begin{table}[t]
	\centering
	\setlength{\tabcolsep}{1.14mm}
	\caption{Comparisons of training time, test time and computational memory on ModelNet \cite{7298801} dataset. }
	\scalebox{0.90}{
		\begin{tabular}{|c|ccc|}
			\hline
			Comparison Methods & Training Time (h) & Test Time (s) & \#Parameters (M) \\
			\hline
			iCaRL \cite{Rebuffi_2017_CVPR} & 1.16 & 0.15 & 3.34 \\
			EEIL \cite{Castro_2018_ECCV} & 1.19 & 0.14 & 3.35 \\
			IL2M \cite{Belouadah_2019_ICCV} & 1.19 & 0.14 & 3.34 \\
			DGMa \cite{Oleksiy_2019_CVPR} & 4.13 & 0.29 & 6.73 \\
			BiC \cite{Wu_2019_CVPR} & 1.72 & 0.14 & 3.34 \\
			RPS-Net \cite{NIPS2019_9429} & 1.45 & 0.21 & 7.68 \\
			LUCIR \cite{Hou_2019_CVPR} + DDE \cite{Hu_20121_CVPR} & 1.64 & 0.22 & 3.34 \\
			LUCIR \cite{Hou_2019_CVPR} + GeoDL \cite{Christian2021MGeoCont} & 1.85 & 0.25 & 3.34 \\
			I3DOL \cite{I3DOL_Dong_AAAI2021} & 1.26 & 0.15 & 3.45 \\
			Ours & 1.41 & 0.17 & 3.65 \\
			\hline 	
		\end{tabular}
	} 	
	\label{tab:time_memory_comparison}
\end{table}

\begin{table}[t]
	\centering
	\setlength{\tabcolsep}{1.1mm}
	\caption{Investigations of catastrophic forgetting on ModelNet \cite{7298801} dataset in terms of F1 score and Recall. }
	\scalebox{0.999}{
		\begin{tabular}{|c|cc|cc|}
			\hline
			Comparison Methods & F1 Score (\%) & $\triangle$ (\%) & Recall (\%) & $\triangle$ (\%) \\
			\hline
			iCaRL \cite{Rebuffi_2017_CVPR} & 69.4 & $\downarrow$19.7 & 66.5 & $\downarrow$18.3 \\
			EEIL \cite{Castro_2018_ECCV} & 77.3 & $\downarrow$11.8 & 74.1 & $\downarrow$10.7 \\
			IL2M \cite{Belouadah_2019_ICCV} & 82.9 & $\downarrow$6.2 & 79.7 & $\downarrow$5.1 \\
			DGMa \cite{Oleksiy_2019_CVPR} & 78.4 & $\downarrow$10.7 & 73.8 & $\downarrow$11.0 \\
			BiC \cite{Wu_2019_CVPR} & 83.5 & $\downarrow$5.6 & 78.5 & $\downarrow$6.3  \\
			RPS-Net \cite{NIPS2019_9429} & 81.3 & $\downarrow$7.8 & 80.3 & $\downarrow$4.5  \\
			LUCIR \cite{Hou_2019_CVPR} + DDE \cite{Hu_20121_CVPR} & 85.1 & $\downarrow$4.0 & 80.6 & $\downarrow$4.2 \\
			LUCIR \cite{Hou_2019_CVPR} + GeoDL \cite{Christian2021MGeoCont} & 85.4 & $\downarrow$3.7 & 81.5 & $\downarrow$3.3 \\
			I3DOL \cite{I3DOL_Dong_AAAI2021} & 86.8 & $\downarrow$2.3 & 82.1 & $\downarrow$2.7 \\
			Ours & \textcolor[rgb]{0.698,0.133,0.133}{\textbf{89.1}} & ---& \textcolor[rgb]{0.698,0.133,0.133}{\textbf{84.8}} & --- \\
			\hline 	
		\end{tabular}
	} 	
	\label{tab:computation_f1_recall}
\end{table}

\subsection{Qualitative Analysis of Catastrophic Forgetting}
In this subsection, we present the comparison experiments on ModelNet \cite{7298801} dataset in terms of averaged F1 score and Recall across all incremental states to investigate catastrophic forgetting, as shown in Table~\ref{tab:computation_f1_recall}. Our proposed model outperforms other baseline comparison methods \cite{Rebuffi_2017_CVPR, Castro_2018_ECCV, Belouadah_2019_ICCV, Wu_2019_CVPR, Oleksiy_2019_CVPR, Hu_20121_CVPR, Christian2021MGeoCont, I3DOL_Dong_AAAI2021} about 2.3\%$\sim$19.7\% in terms of averaged F1 score and Recall. This improvement validates the superiority of our InOR-Net model to identify new categories consecutively under a streaming manner. Moreover, it verifies our model could perform well across all classes (both old and new classes) rather than only some specific classes, and also illustrates the effectiveness of our model to alleviate catastrophic forgetting on old classes.

\begin{table}[t]
	\centering
	\setlength{\tabcolsep}{2.2mm}
	\caption{Comparisons of $p$ value of t-test on benchmark datasets.}
	\scalebox{0.8}{
		\begin{tabular}{|c|ccc|}
			\hline
			Variants & ShapeNet \cite{DBLP:journals/corr/ChangFGHHLSSSSX15} & ModelNet \cite{7298801} & ScanNet \cite{8099744}  \\
			\hline
			Ours vs IL2M \cite{Belouadah_2019_ICCV} & $7.05\times10^{-5}$ & $6.12\times10^{-4}$ & $5.58\times10^{-4}$  \\
			DGMw \cite{Oleksiy_2019_CVPR} & $5.13\times10^{-4}$ & $3.72\times10^{-5}$ & $4.21\times10^{-4}$ \\
			DGMa \cite{Oleksiy_2019_CVPR} & $9.17\times10^{-3}$ & $5.17\times10^{-5}$ & $6.48\times10^{-3}$ \\
			Ours vs BiC \cite{Wu_2019_CVPR} & $1.18\times10^{-5}$ & $8.62\times10^{-4}$ & $7.15\times10^{-4}$ \\
			Ours vs RPS-Net \cite{NIPS2019_9429} & $5.61\times10^{-3}$ & $7.44\times10^{-5}$ & $8.93\times10^{-5}$ \\
			
			Ours vs LUCIR \cite{Hou_2019_CVPR} + DDE \cite{Hu_20121_CVPR} & $2.85\times10^{-4}$ & $5.17\times10^{-4}$ & $8.94\times10^{-5}$ \\
			Ours vs LUCIR \cite{Hou_2019_CVPR} + GeoDL \cite{Christian2021MGeoCont} & $6.29\times10^{-4}$ & $7.33\times10^{-5}$ & $8.21\times10^{-4}$ \\
			Ours vs I3DOL \cite{I3DOL_Dong_AAAI2021} & $9.14\times10^{-5}$ & $8.06\times10^{-5}$ & $7.73\times10^{-5}$ \\ 
			
			\hline
		\end{tabular}
	} 				
	\label{tab:p_value_significant_improvement}
\end{table}

\subsection{Qualitative Analysis of Substantial Improvement}
In order to show whether the improvement of our INOR-Net model compared with other baseline approaches is significant, we introduce the t-test experiments via five random runs in this subsection. When the $p$ value of t-test is lower than 0.05, we consider the improvement of object recognition ability is significant. As presented in Table~\ref{tab:p_value_significant_improvement}, the $p$ value of Ours vs I3DOL \cite{I3DOL_Dong_AAAI2021} is much lower than 0.05, which validates that our InOR-Net model has a substantial improvement over the previous conference version I3DOL \cite{I3DOL_Dong_AAAI2021} on three benchmark datasets. In addition, we introduce t-test experiments between Ours and other baselines \cite{Belouadah_2019_ICCV, Wu_2019_CVPR, NIPS2019_9429, Oleksiy_2019_CVPR, Hu_20121_CVPR, Christian2021MGeoCont}. The $p$ values of these comparisons are significantly lower than 0.05, which supports the superior recognition results of our model to tackle catastrophic forgetting on old categories of 3D objects.

\section{Conclusion and Future Work}
In this paper, we propose a novel Incremental 3D Object Recognition Network (InOR-Net) to recognize novel categories of 3D objects under a streaming manner, without the catastrophic forgetting on old 3D classes. Specifically, we focus on capturing the distinctive 3D characteristics within each class via a category-guided geometric reasoning, and identifying which 3D geometric characteristics are important to alleviate the forgetting on old classes of 3D objects via a critic-induced geometric attention. A dual adaptive fairness compensations strategy including both weights and prediction scores corrections is designed to tackle the forgetting brought by class imbalance problem. We verify the superior performance of the proposed InOR-Net model via extensive comparison experiments on representative point cloud datasets.

The proposed model lacks some theoretical analysis to guarantee the convergence in the theory perspective, and may suffer from recognition performance degradation when the collected point cloud is heavily polluted by noise and loses some core geometric structures. 
In the future, we will focus on addressing these limitations, and extend our proposed model into several challenging 3D vision fields such as intelligent robotics and scene understanding.

\ifCLASSOPTIONcaptionsoff
  \newpage
\fi

\bibliographystyle{IEEEtran}
\bibliography{Arxiv}

\begin{thebibliography}{10}
\providecommand{\url}[1]{#1}
\csname url@samestyle\endcsname
\providecommand{\newblock}{\relax}
\providecommand{\bibinfo}[2]{#2}
\providecommand{\BIBentrySTDinterwordspacing}{\spaceskip=0pt\relax}
\providecommand{\BIBentryALTinterwordstretchfactor}{4}
\providecommand{\BIBentryALTinterwordspacing}{\spaceskip=\fontdimen2\font plus
\BIBentryALTinterwordstretchfactor\fontdimen3\font minus
  \fontdimen4\font\relax}
\providecommand{\BIBforeignlanguage}[2]{{%
\expandafter\ifx\csname l@#1\endcsname\relax
\typeout{** WARNING: IEEEtran.bst: No hyphenation pattern has been}%
\typeout{** loaded for the language `#1'. Using the pattern for}%
\typeout{** the default language instead.}%
\else
\language=\csname l@#1\endcsname
\fi
#2}}
\providecommand{\BIBdecl}{\relax}
\BIBdecl

\bibitem{8593741}
J.~{Stria} and V.~{Hlavác}, ``Classification of hanging garments using learned
  features extracted from 3d point clouds,'' in \emph{2018 IEEE/RSJ
  International Conference on Intelligent Robots and Systems (IROS)}, 2018, pp.
  5307--5312.

\bibitem{Roynard_2018_CVPR_Workshops}
X.~Roynard, J.-E. Deschaud, and F.~Goulette, ``Paris-lille-3d: A point cloud
  dataset for urban scene segmentation and classification,'' in \emph{The IEEE
  Conference on Computer Vision and Pattern Recognition (CVPR) Workshops}, June
  2018, pp. 2108--21\,083.

\bibitem{Behl_2017_ICCV}
A.~Behl, O.~Hosseini~Jafari, S.~Karthik~Mustikovela, H.~Abu~Alhaija, C.~Rother,
  and A.~Geiger, ``Bounding boxes, segmentations and object coordinates: How
  important is recognition for 3d scene flow estimation in autonomous driving
  scenarios?'' in \emph{The IEEE International Conference on Computer Vision
  (ICCV)}, Oct 2017, pp. 2593--2602.

\bibitem{What_Transferred_Dong_CVPR2020}
J.~Dong, Y.~Cong, G.~Sun, B.~Zhong, and X.~Xu, ``What can be transferred:
  Unsupervised domain adaptation for endoscopic lesions segmentation,'' in
  \emph{IEEE/CVF Conference on Computer Vision and Pattern Recognition (CVPR)},
  June 2020, pp. 4022--4031.

\bibitem{Qi_2017_CVPR}
C.~R. Qi, H.~Su, K.~Mo, and L.~J. Guibas, ``Pointnet: Deep learning on point
  sets for 3d classification and segmentation,'' in \emph{The IEEE Conference
  on Computer Vision and Pattern Recognition (CVPR)}, July 2017, pp. 77--85.

\bibitem{NIPS2017_7095}
C.~R. Qi, L.~Yi, H.~Su, and L.~J. Guibas, ``Pointnet++: Deep hierarchical
  feature learning on point sets in a metric space,'' in \emph{Advances in
  Neural Information Processing Systems 30}, I.~Guyon, U.~V. Luxburg,
  S.~Bengio, H.~Wallach, R.~Fergus, S.~Vishwanathan, and R.~Garnett, Eds.\hskip
  1em plus 0.5em minus 0.4em\relax Curran Associates, Inc., 2017, pp.
  5099--5108.

\bibitem{NIPS2018_7362}
Y.~Li, R.~Bu, M.~Sun, W.~Wu, X.~Di, and B.~Chen, ``Pointcnn: Convolution on
  x-transformed points,'' in \emph{Advances in Neural Information Processing
  Systems 31}, S.~Bengio, H.~Wallach, H.~Larochelle, K.~Grauman,
  N.~Cesa-Bianchi, and R.~Garnett, Eds.\hskip 1em plus 0.5em minus 0.4em\relax
  Curran Associates, Inc., 2018, pp. 820--830.

\bibitem{Lan_2019_CVPR}
S.~Lan, R.~Yu, G.~Yu, and L.~S. Davis, ``Modeling local geometric structure of
  3d point clouds using geo-cnn,'' in \emph{Proceedings of the IEEE/CVF
  Conference on Computer Vision and Pattern Recognition (CVPR)}, June 2019, pp.
  998--1008.

\bibitem{yan2020pointasnl}
X.~Yan, C.~Zheng, Z.~Li, S.~Wang, and S.~Cui, ``Pointasnl: Robust point clouds
  processing using nonlocal neural networks with adaptive sampling,'' in
  \emph{Proceedings of the IEEE/CVF Conference on Computer Vision and Pattern
  Recognition}, 2020, pp. 5589--5598.

\bibitem{yu2022generalized}
W.~Yu, H.~Yu, Y.~Huang, and L.~Wang, ``Generalized inter-class loss for gait
  recognition,'' in \emph{Proceedings of the 28th ACM international conference
  on multimedia}, 2022, pp. 141--150.

\bibitem{10.1007/978-3-319-46493-0_37}
Z.~Li and D.~Hoiem, ``Learning without forgetting,'' in \emph{Computer Vision
  -- ECCV 2016}, B.~Leibe, J.~Matas, N.~Sebe, and M.~Welling, Eds.\hskip 1em
  plus 0.5em minus 0.4em\relax Cham: Springer International Publishing, 2016,
  pp. 614--629.

\bibitem{Rebuffi_2017_CVPR}
S.-A. Rebuffi, A.~Kolesnikov, G.~Sperl, and C.~H. Lampert, ``icarl: Incremental
  classifier and representation learning,'' in \emph{The IEEE Conference on
  Computer Vision and Pattern Recognition (CVPR)}, July 2017, pp. 5533--5542.

\bibitem{dong2022federated}
J.~Dong, L.~Wang, Z.~Fang, G.~Sun, S.~Xu, X.~Wang, and Q.~Zhu, ``Federated
  class-incremental learning,'' in \emph{IEEE/CVF Conference on Computer Vision
  and Pattern Recognition (CVPR)}, June 2022, pp. 10\,164--10\,173.

\bibitem{wei2020lifelong}
K.~Wei, C.~Deng, and X.~Yang, ``Lifelong zero-shot learning.'' in \emph{IJCAI},
  2020, pp. 551--557.

\bibitem{Castro_2018_ECCV}
F.~M. Castro, M.~J. Marin-Jimenez, N.~Guil, C.~Schmid, and K.~Alahari,
  ``End-to-end incremental learning,'' in \emph{The European Conference on
  Computer Vision (ECCV)}, September 2018, pp. 241--257.

\bibitem{Wu_2019_CVPR}
Y.~Wu, Y.~Chen, L.~Wang, Y.~Ye, Z.~Liu, Y.~Guo, and Y.~Fu, ``Large scale
  incremental learning,'' in \emph{The IEEE Conference on Computer Vision and
  Pattern Recognition (CVPR)}, June 2019, pp. 374--382.

\bibitem{8674766}
Z.~Yang, S.~Al-Dahidi, P.~Baraldi, E.~Zio, and L.~Montelatici, ``A novel
  concept drift detection method for incremental learning in nonstationary
  environments,'' \emph{IEEE Transactions on Neural Networks and Learning
  Systems}, vol.~31, no.~1, pp. 309--320, 2020.

\bibitem{I3DOL_Dong_AAAI2021}
J.~Dong, Y.~Cong, G.~Sun, B.~Ma, and L.~Wang, ``I3dol: Incremental 3d object
  learning without catastrophic forgetting,'' \emph{Proceedings of the AAAI
  Conference on Artificial Intelligence}, vol.~35, no.~7, pp. 6066--6074, May
  2021.

\bibitem{Li_2018_CVPR}
J.~Li, B.~M. Chen, and G.~Hee~Lee, ``So-net: Self-organizing network for point
  cloud analysis,'' in \emph{The IEEE Conference on Computer Vision and Pattern
  Recognition (CVPR)}, June 2018, pp. 9397--9406.

\bibitem{Oleksiy_2019_CVPR}
O.~Ostapenko, M.~M. Puscas, T.~Klein, P.~J{\"{a}}hnichen, and M.~Nabi,
  ``Learning to remember: {A} synaptic plasticity driven framework for
  continual learning,'' in \emph{The IEEE Conference on Computer Vision and
  Pattern Recognition (CVPR)}, June 2019, p. 11321–11329.

\bibitem{wei2021incremental222}
K.~Wei, C.~Deng, X.~Yang, and M.~Li, ``Incremental embedding learning via
  zero-shot translation,'' in \emph{Proceedings of the AAAI Conference on
  Artificial Intelligence}, vol.~35, no.~11, 2021, pp. 10\,254--10\,262.

\bibitem{chen2019two}
C.~Chen, H.~Wang, W.~Liu, X.~Zhao, T.~Hu, and G.~Chen, ``Two-stage label
  embedding via neural factorization machine for multi-label classification,''
  in \emph{Proceedings of the AAAI Conference on Artificial Intelligence},
  vol.~33, no.~01, 2019, pp. 3304--3311.

\bibitem{Shmelkov_2017_ICCV}
K.~Shmelkov, C.~Schmid, and K.~Alahari, ``Incremental learning of object
  detectors without catastrophic forgetting,'' in \emph{The IEEE International
  Conference on Computer Vision (ICCV)}, Oct 2017, pp. 3400--3409.

\bibitem{yu2022cyclic}
H.~Yu, H.~Peng, Y.~Huang, J.~Fu, H.~Du, L.~Wang, and H.~Ling, ``Cyclic
  differentiable architecture search,'' \emph{IEEE Transactions on Pattern
  Analysis and Machine Intelligence}, vol.~45, no.~1, pp. 211--228, 2022.

\bibitem{9459438}
J.~Leo and J.~Kalita, ``Incremental deep neural network learning using
  classification confidence thresholding,'' \emph{IEEE Transactions on Neural
  Networks and Learning Systems}, vol.~33, no.~12, pp. 7706--7716, 2022.

\bibitem{9877899}
K.~Wei, D.~Chen, Y.~Li, X.~Yang, C.~Deng, and D.~Tao, ``Incremental embedding
  learning with disentangled representation translation,'' \emph{IEEE
  Transactions on Neural Networks and Learning Systems}, vol.~1, no.~2, pp.
  1--13, 2022.

\bibitem{9737321}
L.~Yu, X.~Liu, and J.~van~de Weijer, ``Self-training for class-incremental
  semantic segmentation,'' \emph{IEEE Transactions on Neural Networks and
  Learning Systems}, vol.~1, no.~1, pp. 1--12, 2022.

\bibitem{9533187}
D.-W. Zhou, Y.~Yang, and D.-C. Zhan, ``Learning to classify with incremental
  new class,'' \emph{IEEE Transactions on Neural Networks and Learning
  Systems}, vol.~33, no.~6, pp. 2429--2443, 2022.

\bibitem{Kirkpatrick3521}
J.~Kirkpatrick, R.~Pascanu, N.~Rabinowitz, J.~Veness, G.~Desjardins, A.~A.
  Rusu, K.~Milan, J.~Quan, T.~Ramalho, A.~Grabska-Barwinska, D.~Hassabis,
  C.~Clopath, D.~Kumaran, and R.~Hadsell, ``Overcoming catastrophic forgetting
  in neural networks,'' \emph{Proceedings of the National Academy of Sciences},
  vol. 114, no.~13, pp. 3521--3526, 2017.

\bibitem{9616392_Dong}
J.~Dong, Y.~Cong, G.~Sun, Z.~Fang, and Z.~Ding, ``Where and how to transfer:
  Knowledge aggregation-induced transferability perception for unsupervised
  domain adaptation,'' \emph{IEEE Transactions on Pattern Analysis and Machine
  Intelligence}, pp. 1--1, 2021.

\bibitem{NIPS2018_7836}
C.~Wu, L.~Herranz, X.~Liu, y.~wang, J.~van~de Weijer, and B.~Raducanu, ``Memory
  replay gans: Learning to generate new categories without forgetting,'' in
  \emph{Advances in Neural Information Processing Systems 31}, S.~Bengio,
  H.~Wallach, H.~Larochelle, K.~Grauman, N.~Cesa-Bianchi, and R.~Garnett,
  Eds.\hskip 1em plus 0.5em minus 0.4em\relax Curran Associates, Inc., 2018,
  pp. 5962--5972.

\bibitem{deesil-eccv2018}
E.~Belouadah and A.~Popescu, ``{DeeSIL:} deep-shallow incremental learning,''
  \emph{TaskCV Workshop @ ECCV 2018.}, p. 151–157, 2018.

\bibitem{Belouadah_2019_ICCV}
E.~Belouadah, A.~Popescu, and F.~M. Castro, ``Il2m: Class incremental learning
  with dual memory,'' in \emph{2019 IEEE/CVF International Conference on
  Computer Vision (ICCV)}, 2019, pp. 583--592.

\bibitem{DBLP:journals/corr/RusuRDSKKPH16}
A.~A. Rusu, N.~C. Rabinowitz, G.~Desjardins, H.~Soyer, J.~Kirkpatrick,
  K.~Kavukcuoglu, R.~Pascanu, and R.~Hadsell, ``Progressive neural networks,''
  \emph{arXiv preprint arXiv:1606.04671}, 2016.

\bibitem{NIPS2019_9429}
J.~Rajasegaran, M.~Hayat, S.~H. Khan, F.~S. Khan, and L.~Shao, ``Random path
  selection for continual learning,'' in \emph{Advances in Neural Information
  Processing Systems 32}.\hskip 1em plus 0.5em minus 0.4em\relax Curran
  Associates, Inc., 2019, pp. 12\,669--12\,679.

\bibitem{Christian2021MGeoCont}
C.~Simon, P.~Koniusz, and M.~Harandi, ``On learning the geodesic path for
  incremental learning,'' in \emph{Proceedings of the IEEE/CVF Conference on
  Computer Vision and Pattern Recognition (CVPR)}, 2021, pp. 1591--1600.

\bibitem{Hu_20121_CVPR}
X.~Hu, K.~Tang, C.~Miao, X.-S. Hua, and H.~Zhang, ``Distilling causal effect of
  data in class-incremental learning,'' in \emph{The IEEE Conference on
  Computer Vision and Pattern Recognition (CVPR)}, June 2021, pp. 3957--3966.

\bibitem{DBLP:journals/corr/DaiQXLZHW17}
J.~Dai, H.~Qi, Y.~Xiong, Y.~Li, G.~Zhang, H.~Hu, and Y.~Wei, ``Deformable
  convolutional networks,'' \emph{arXiv preprint arXiv:1409.7495}, 2017.

\bibitem{NIPS2019_8940}
C.~Qin, H.~You, L.~Wang, C.-C.~J. Kuo, and Y.~Fu, ``Pointdan: A multi-scale 3d
  domain adaption network for point cloud representation,'' in \emph{Advances
  in Neural Information Processing Systems 32}.\hskip 1em plus 0.5em minus
  0.4em\relax Curran Associates, Inc., 2019, pp. 7190--7201.

\bibitem{1053964}
T.~Cover and P.~Hart, ``Nearest neighbor pattern classification,'' \emph{IEEE
  Transactions on Information Theory}, vol.~13, no.~1, pp. 21--27, 1967.

\bibitem{10.1007/978-3-540-74976-9_25}
Y.~Song, J.~Huang, D.~Zhou, H.~Zha, and C.~L. Giles, ``Iknn: Informative
  k-nearest neighbor pattern classification,'' in \emph{Knowledge Discovery in
  Databases: PKDD 2007}, 2007, pp. 248--264.

\bibitem{yang2020heterogeneous}
X.~Yang, C.~Deng, T.~Liu, and D.~Tao, ``Heterogeneous graph attention network
  for unsupervised multiple-target domain adaptation,'' \emph{IEEE Transactions
  on Pattern Analysis and Machine Intelligence}, vol.~44, no.~4, pp.
  1992--2003, 2020.

\bibitem{zhang2021practical}
J.~Zhang, C.~Chen, B.~Li, L.~Lyu, S.~Wu, J.~Xu, S.~Ding, and C.~Wu, ``A
  practical data-free approach to one-shot federated learning with
  heterogeneity,'' \emph{arXiv preprint arXiv:2112.12371}, 2021.

\bibitem{shen2020federated}
T.~Shen, J.~Zhang, X.~Jia, F.~Zhang, G.~Huang, P.~Zhou, K.~Kuang, F.~Wu, and
  C.~Wu, ``Federated mutual learning,'' \emph{arXiv preprint arXiv:2006.16765},
  2020.

\bibitem{9449988}
J.~Deng, J.~Guo, J.~Yang, N.~Xue, I.~Cotsia, and S.~P. Zafeiriou, ``Arcface:
  Additive angular margin loss for deep face recognition,'' \emph{IEEE
  Transactions on Pattern Analysis and Machine Intelligence}, vol.~1, no.~1,
  pp. 4685--4694, 2021.

\bibitem{Liu_2017_CVPR}
W.~Liu, Y.~Wen, Z.~Yu, M.~Li, B.~Raj, and L.~Song, ``Sphereface: Deep
  hypersphere embedding for face recognition,'' in \emph{The IEEE Conference on
  Computer Vision and Pattern Recognition (CVPR)}, 2017, pp. 212--220.

\bibitem{fu2019dual}
J.~Fu, J.~Liu, H.~Tian, Y.~Li, Y.~Bao, Z.~Fang, and H.~Lu, ``Dual attention
  network for scene segmentation,'' in \emph{Proceedings of the IEEE Conference
  on Computer Vision and Pattern Recognition}, 2019, pp. 3146--3154.

\bibitem{DBLP:journals/corr/ChangFGHHLSSSSX15}
A.~X. Chang, T.~A. Funkhouser, L.~J. Guibas, P.~Hanrahan, Q.~Huang, Z.~Li,
  S.~Savarese, M.~Savva, S.~Song, H.~Su, J.~Xiao, L.~Yi, and F.~Yu, ``Shapenet:
  An information-rich 3d model repository,'' \emph{arXiv preprint
  arXiv:1512.03012}, 2015.

\bibitem{7298801}
{Zhirong Wu}, S.~{Song}, A.~{Khosla}, {Fisher Yu}, {Linguang Zhang}, {Xiaoou
  Tang}, and J.~{Xiao}, ``3d shapenets: A deep representation for volumetric
  shapes,'' in \emph{2015 IEEE Conference on Computer Vision and Pattern
  Recognition (CVPR)}, 2015, pp. 1912--1920.

\bibitem{8099744}
A.~{Dai}, A.~X. {Chang}, M.~{Savva}, M.~{Halber}, T.~{Funkhouser}, and
  M.~{Nießner}, ``Scannet: Richly-annotated 3d reconstructions of indoor
  scenes,'' in \emph{2017 IEEE Conference on Computer Vision and Pattern
  Recognition (CVPR)}, 2017, pp. 2432--2443.

\bibitem{Hou_2019_CVPR}
S.~Hou, X.~Pan, C.~C. Loy, Z.~Wang, and D.~Lin, ``Learning a unified classifier
  incrementally via rebalancing,'' in \emph{The IEEE Conference on Computer
  Vision and Pattern Recognition (CVPR)}, June 2019, pp. 831--839.

\bibitem{Ceri2013}
S.~Ceri, A.~Bozzon, M.~Brambilla, E.~Della~Valle, P.~Fraternali, and
  S.~Quarteroni, \emph{An Introduction to Information Retrieval}.\hskip 1em
  plus 0.5em minus 0.4em\relax Berlin, Heidelberg: Springer Berlin Heidelberg,
  2013, pp. 3--11.

\bibitem{wei2021incremental}
K.~Wei, C.~Deng, X.~Yang, and D.~Tao, ``Incremental zero-shot learning,''
  \emph{IEEE Transactions on Cybernetics}, vol.~52, no.~12, pp.
  13\,788--13\,799, 2022.

\end{thebibliography}

\begin{IEEEbiography}[{\includegraphics[width=1in,height=1.25in,clip,keepaspectratio]{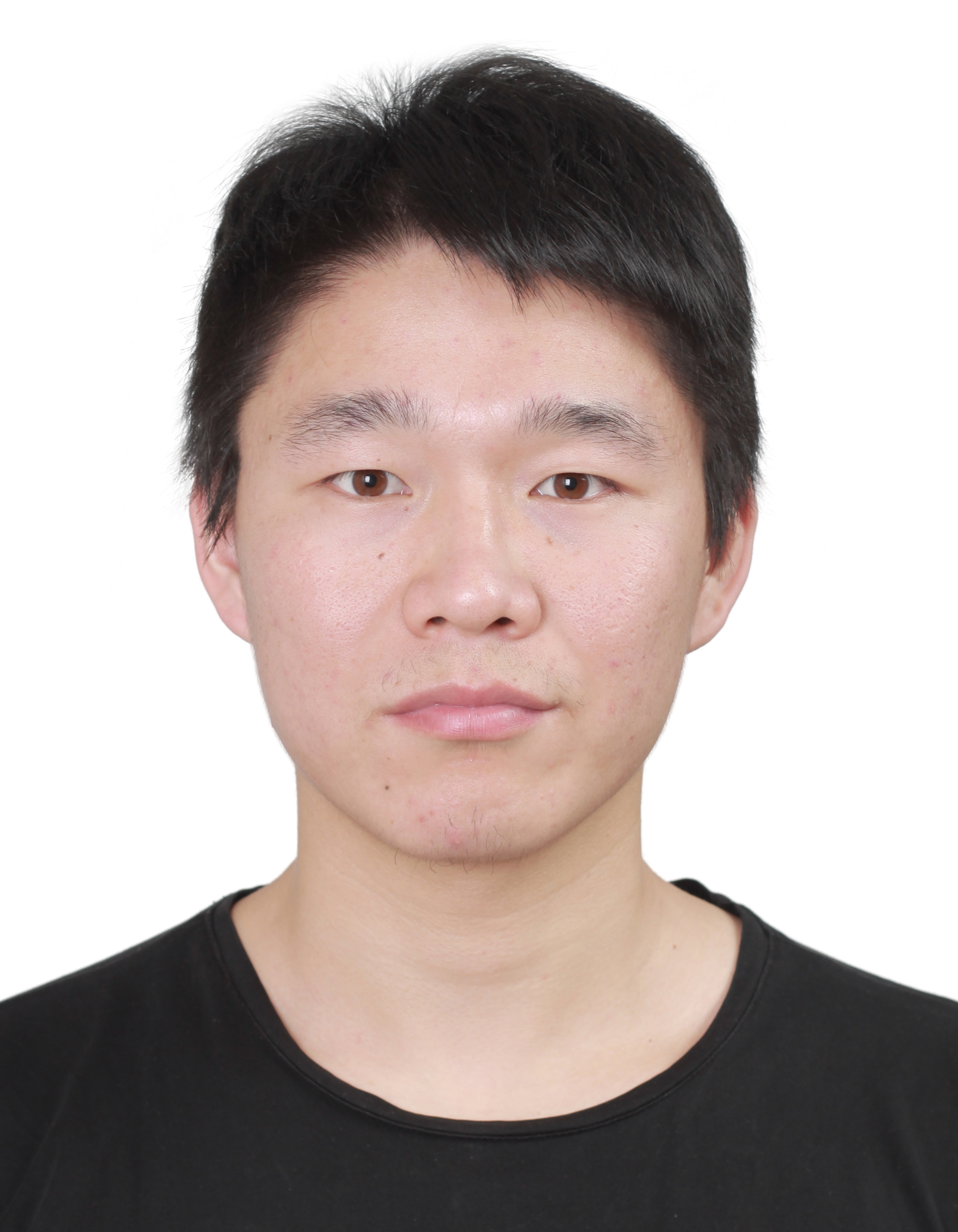}}]{Jiahua Dong} is currently a PhD candidate in the State Key Laboratory of Robotics, Shenyang Institute of Automation, Chinese Academy of Sciences. He received the B.S. degree from Jilin University in 2017, and visited ETH Z\"{u}rich from April 2022 to September 2022, Max Planck Institute for Informatics from October 2022 to January 2023. His current research interests include transfer learning, class-incremental learning, federated learning and medical image analysis.
\end{IEEEbiography}

\begin{IEEEbiography}[{\includegraphics[width=1in,height=1.25in,clip,keepaspectratio]{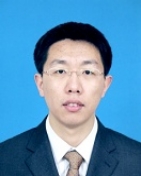}}]{Yang Cong} (S’09-M’11-SM’15) is a full professor of Chinese Academy of Sciences. He received the B.S. degree from Northeast University in 2004, and the Ph.D. degree from State Key Laboratory of Robotics, Chinese Academy of Sciences in 2009. He was a Research Fellow of National University of Singapore (NUS) and Nanyang Technological University (NTU) from 2009 to 2011, respectively; and a visiting scholar of University of Rochester. His current research interests include robot vision, robot learning, multimedia and medical image analysis. 
\end{IEEEbiography}

\begin{IEEEbiography}[{\includegraphics[width=1in,height=1.25in,clip,keepaspectratio]{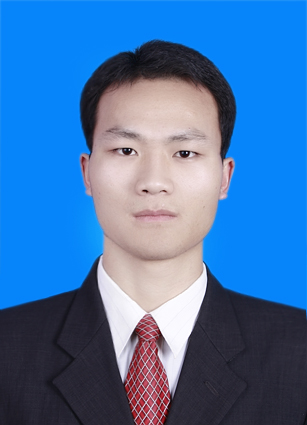}}]{Gan Sun} (S’19-M’20) is an associate professor in Shenyang Institute of Automation, Chinese Academy of Sciences. He received the B.S. degree from Shandong Agricultural University in 2013, the Ph.D. degree from Shenyang Institute of Automation, Chinese Academy of Sciences in 2020, and visited Northeastern University from April 2018 to May 2019, Massachusetts Institute of Technology from June 2019 to November 2019. His current research interests include lifelong learning, incremental learning and medical data analysis.
\end{IEEEbiography}

\begin{IEEEbiography}[{\includegraphics[width=1in,height=1.25in,clip,keepaspectratio]{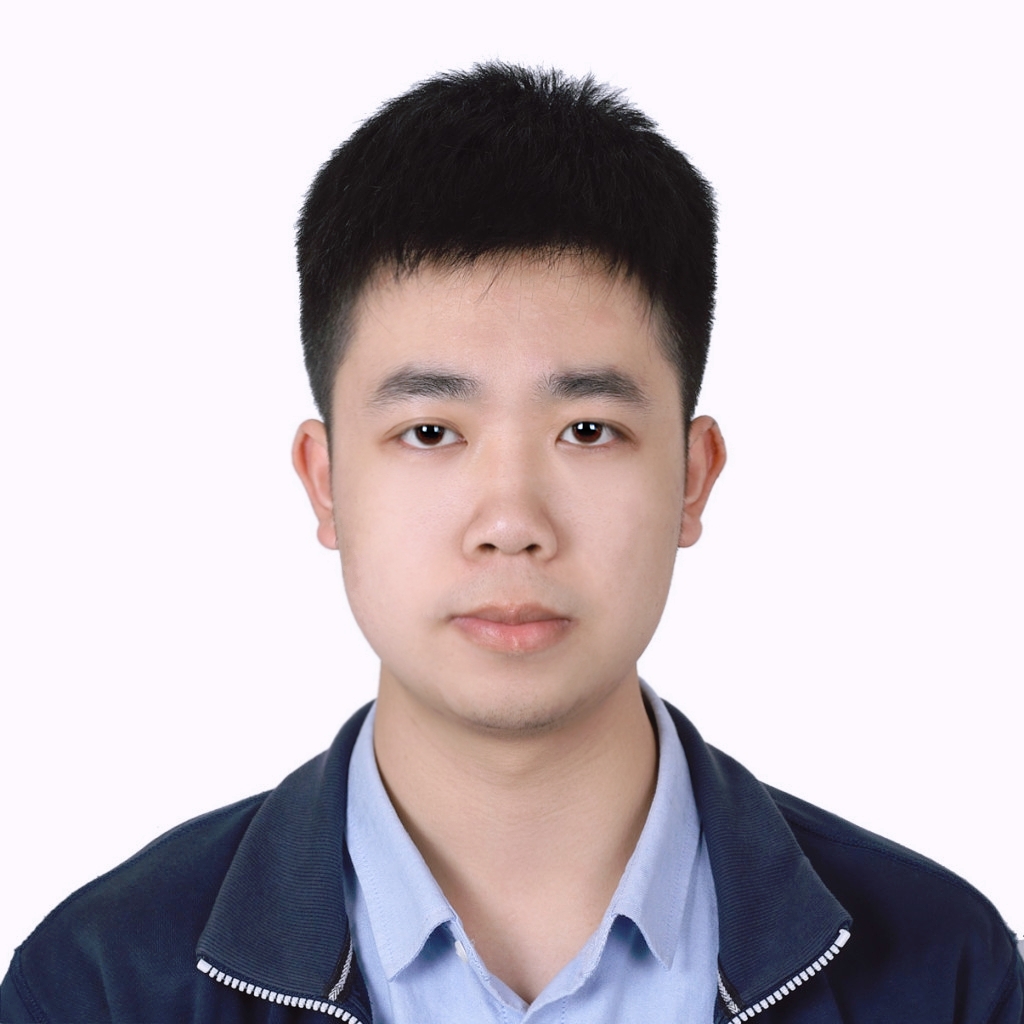}}]{Lixu Wang} is currently a first-year Ph.D student at Northwestern University, U.S. He received the B.E. degree from Zhejiang University, China, in 2020. His current research interests include unsupervised learning, neural network robustness and AI security.
\end{IEEEbiography}

\begin{IEEEbiography}[{\includegraphics[width=1in,height=1.25in,clip,keepaspectratio]{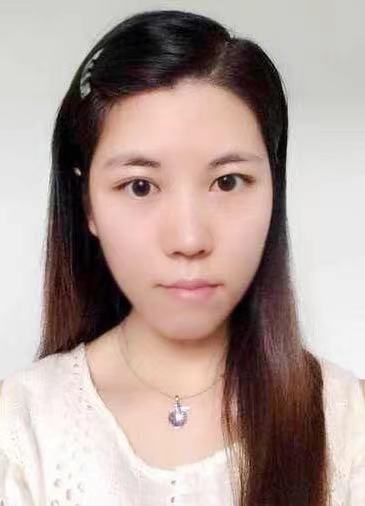}}]{Lingjuan Lyu} is currently a senior research scientist and team leader in Sony AI. She received her PhD degree from the University of Melbourne in 2018. She was a winner of the IBM Fellowship program (50 winners Worldwide) and contributed to various professional activities. Her current research interest is trustworthy AI. She has publications in NeurIPS, ICLR, AAAI, IJCAI, etc. Her paper won several best paper awards in top venues. 
\end{IEEEbiography}

\begin{IEEEbiography}[{\includegraphics[width=1in,height=1.25in,clip,keepaspectratio]{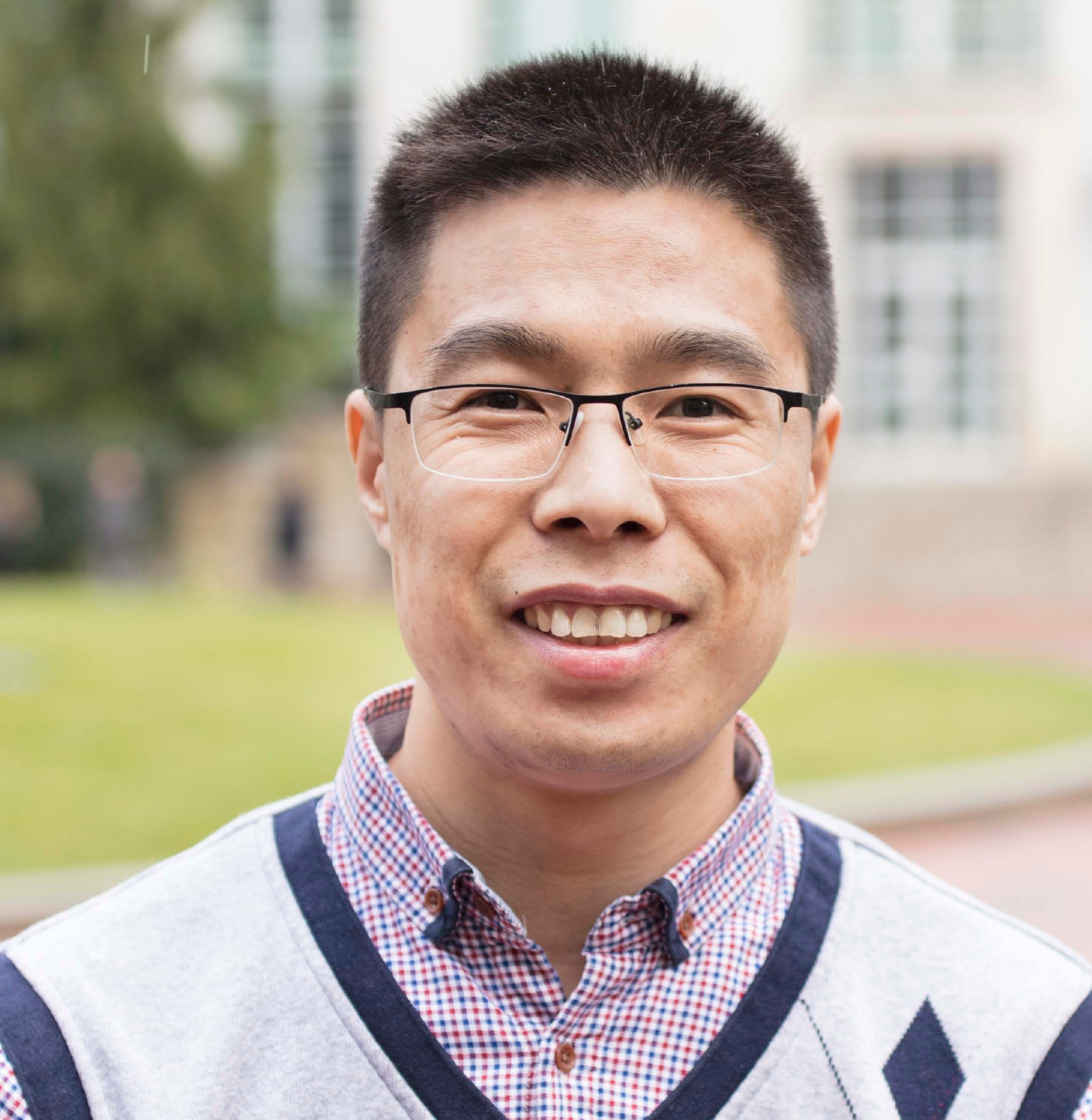}}]{Jun Li} (M'16) received the B.A. from Pan Zhi Hua University in 2006. He received the M.S. from China West Normal University in 2009 and the PhD degree from the Nanjing University of Science and Technology in 2015. From Oct. 2012 to July 2013, he was a visiting student in Rutgers University, Piscataway, NJ, USA. From Dec. 2015 to Oct. 2018, he was a postdoctoral associate with the Northeastern University, Boston, MA, USA. From Nov. 2018 to Oct. 2019, he was a postdoctoral associate with the Institute of Medical Engineering and Science, Massachusetts Institute of Technology, Cambridge, MA, USA. He is currently a professor of Nanjing University of Science and Technology, Nanjing, China. His research interests are machine learning and computer vision.
\end{IEEEbiography}

\begin{IEEEbiography}[{\includegraphics[width=1in,height=1.25in,clip,keepaspectratio]{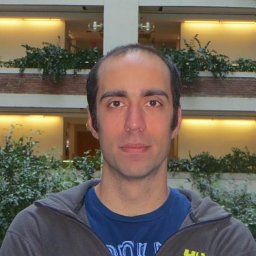}}]{Ender Konukoglu} is currently a Professor of Biomedical Image Computing at ETH-Z\"{u}rich. He received his B.S. and M.S. degrees at Bogazici University in 2003 and 2005. He got the PhD from University of Nice Sophia Antipolis working at INRIA Sophia Antipolis Mediterranean in 2009. After the PhD he worked as a post-doctoral researcher at Microsoft Research in Cambridge between 2009 and 2012. His research interests are computer vision and medical imaging analysis. 
\end{IEEEbiography} 

\end{document}